%% file: acl_latex.tex
\documentclass[11pt]{article}

\usepackage[final]{acl}

\usepackage{times}
\usepackage{latexsym}

\usepackage[T1]{fontenc}

\usepackage[utf8]{inputenc}

\usepackage{microtype}

\usepackage{inconsolata}

\usepackage{graphicx}
\usepackage{hyperref}
\usepackage{amsmath}
\usepackage{booktabs}
\usepackage{multirow}
\usepackage{tabularx}
\usepackage{float}
\usepackage{placeins}

\usepackage[table]{xcolor} 
\usepackage{bold-extra}    

\definecolor{LightGray}{gray}{0.92}        
\definecolor{LightBlue}{rgb}{0.9, 0.95, 1.0} 
\definecolor{DarkGreen}{rgb}{0.0, 0.5, 0.0}  
\definecolor{DarkRed}{rgb}{0.8, 0.0, 0.0}

\newcommand{\gcell}{\cellcolor{LightGray}}  
\newcommand{\bcell}{\cellcolor{LightBlue}}  
\newcommand{\gbf}[1]{\textcolor{DarkGreen}{#1}} 
\newcommand{\rbf}[1]{\textcolor{DarkRed}{#1}}   

\NewDocumentCommand{\lyx}
{ mO{} }{\textcolor{orange}{\textsuperscript{\textit{lyx}}\textsf{\textbf{\small[#1]}}}}

\newcommand{\firstarrow}{\ensuremath{\hookrightarrow}}

\title{VFA: Empowering Multilingual MLLMs via Vision-Free Adaptation}

\author{
  \textbf{Yixia Li}\textsuperscript{1}\thanks{Work done during internship at Microsoft Research Asia.}\thanks{Equal contribution.},
   \textbf{Yaqing Shi}\textsuperscript{3}$^\dagger$,
  \textbf{Zhiwen Ruan}\textsuperscript{1},
  \textbf{Dongdong Zhang}\textsuperscript{2},
  \textbf{Lingjie Jiang}\textsuperscript{4}, \\
  \textbf{Shaohan Huang}\textsuperscript{2},
  \textbf{Yun Chen}\textsuperscript{3,5},
  \textbf{Guanhua Chen}\textsuperscript{1}\thanks{Corresponding Author.},
  \textbf{Furu Wei}\textsuperscript{2} \\
  \textsuperscript{1}Southern University of Science and Technology,
  \textsuperscript{2}Microsoft Research Asia  \\
  \textsuperscript{3}Shanghai University of Finance and Economics,
  \textsuperscript{4}Peking University \\
  \textsuperscript{5}MoE Key Laboratory of Interdisciplinary Research of Computation and Economics 
}

\begin{document}
\maketitle

\begin{abstract}
Multimodal large language models have advanced rapidly, yet most remain English-centric, as scaling multilingual multimodal instruction tuning is limited by the scarcity and high cost of high-quality non-English image–text supervision.
Although multilingual text data is abundant, naive textual fine-tuning can disrupt vision-language alignment and induce catastrophic forgetting.
We propose Vision-Free Adaptation (VFA), a framework that decouples multilingual language enhancement from visual alignment by composing complementary task vectors over a shared LLM backbone. Specifically, we fine-tune a base LLM on multilingual text data to derive a multilingual task vector, which is then merged with the vision-aligned task vector of an MLLM.
Experiments on five MLLMs across six multilingual multimodal benchmarks show consistent improvements while preserving both general multimodal and text-only capabilities. Moreover, using less than 2\% of the text data, VFA narrows the gap to the fully multimodal-trained model, demonstrating its data efficiency.
\end{abstract}

\section{Introduction}

\begin{figure}[t!]
    \centering
    \vspace{-10pt}
    \includegraphics[width=0.9\linewidth]{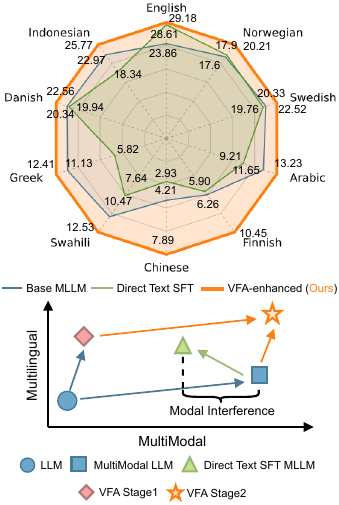}
\vspace{-5pt}
\caption{\textbf{Overview of VFA.} \textbf{Top:} Multilingual visual performance of Idefics3-8B on XM100, where \textsc{VFA} improves multilinguality while direct text-only tuning degrades performance. \textbf{Bottom:} \textsc{VFA} decouples multimodal adaptation via a two-stage vision-free strategy: Stage~1 fine-tunes the LLM on text data, and Stage~2 merges multilingual knowledge into the MLLM.}
    \label{fig:main}
\vspace{-15pt}
\end{figure}

Multimodal large language models (MLLMs) have rapidly advanced the ability to reason over images and text in a unified interface, enabling visual question answering, grounded dialogue, and multimodal reasoning~\citep{tong2025metamorph}. As MLLMs move toward real-world deployment, multilingual and cross-cultural~\citep{song2026culture} competence is becoming increasingly important: models should not only parse and follow non-English instructions, but also ground culturally specific concepts in visual contexts~\citep{bai2025power,nyandwi2025grounding}. Yet most open-source MLLMs remain English-centric, and their performance often degrades substantially when prompts, outputs, or benchmarks shift beyond English, limiting their applicability across global users and scenarios~\citep{yue2024pangea}.

A common approach to building multilingual MLLMs is to scale multilingual multimodal instruction tuning using image–text pairs~\citep{yue2024pangea,dash2025aya}. While effective, this paradigm is fundamentally constrained by data availability: high-quality non-English multimodal supervision remains scarce, expensive to curate, and unevenly distributed across languages and cultures, making comprehensive multilingual coverage difficult to achieve in practice~\citep{zhang2025less}. Moreover, multilingual multimodal training incurs substantial computational overhead, as visual tokens lengthen input sequences and significantly increase attention and memory costs~\citep{kuo2025d}. Taken together, these limitations impede the scalability of large-scale multilingual image–text training as a general solution for multilingual adaptation.

Multilingual text data offers a compelling alternative: it is rich, cheaper to obtain, and far more scalable for multilingual learning. However, directly fine-tuning an MLLM on text can substantially interfere with its established cross-modal alignment, potentially leading to catastrophic forgetting of multimodal capabilities, as shown in Figure~\ref{fig:main}. This failure mode suggests that effective multilingual adaptation should strengthen language competence while minimizing disruption to the vision--language alignment learned during multimodal pre-training~\citep{sanyal2025upweighting}.

To address this challenge, we propose \emph{Vision-Free Adaptation} (VFA), which reframes multilingual and multimodal adaptation as the composition of task vectors defined over a shared LLM backbone. Concretely, we view a trained MLLM as the original LLM augmented by a vision-aligned task vector, representing the parameter changes that encode cross-modal grounding. Separately, we derive a multilingual task vector by fine-tuning the same original LLM on multilingual text data. VFA then fuses these task vectors to inject multilingual competence into the MLLM while preserving its existing vision language alignment. This formulation mitigates the catastrophic forgetting commonly observed in direct text fine-tuning and reduces reliance on scarce multilingual image-text pairs.

We evaluate VFA on five MLLMs across different scales and model families on six multilingual multimodal benchmarks. VFA consistently enhances multilingual performance, yielding improvements of +2.99 and +0.57 on LLaVA-OneVision-1.5 at 8B and 4B scales, respectively.  Importantly, these gains are obtained with general multimodal and text-only capabilities largely retained.
Moreover, despite being trained on only 100K text samples, VFA substantially narrows the gap to multilingual models that rely on millions of image–text pairs for full multimodal training, demonstrating the effectiveness and data efficiency of our approach.
Overall, VFA provides a practical and resource-efficient pathway toward multilingual MLLMs, paving the way for future research on domain-specific multimodal models.\footnote{Code is available at \url{https://github.com/sustech-nlp/VFA}.}

\begin{figure*}[!htbp]
    \centering
    \includegraphics[width=1.0\linewidth]{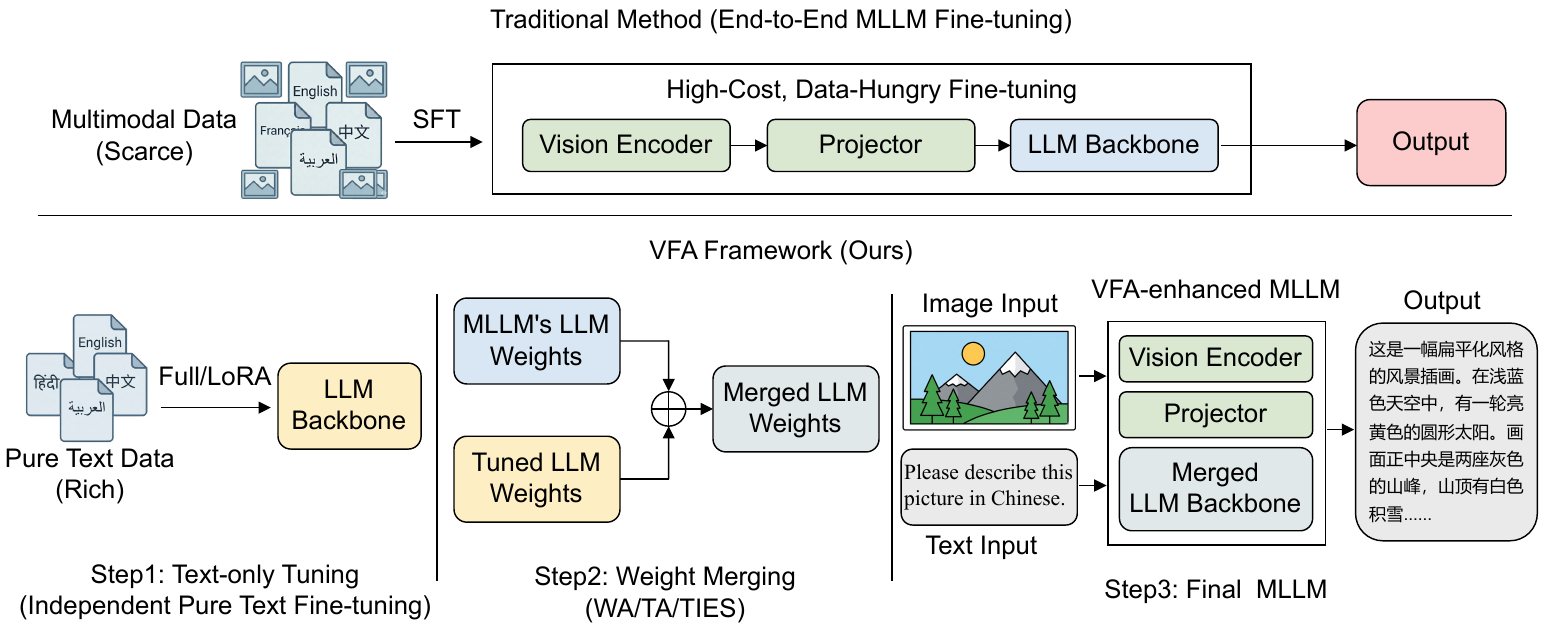}
    \caption{\textbf{Overall Architecture of VFA.} Traditional multilingual MLLM tuning depends on scarce, costly non-English image--text supervision and expensive end-to-end training. \textsc{VFA} leverages rich multilingual \emph{text-only} data, fine-tunes the base LLM to obtain a multilingual task vector, and merges it into a vision-aligned MLLM to enhance multilinguality while preserving vision--language alignment.}
    \label{fig:vfa_framework}
\end{figure*}

\section{Related Work}
    \paragraph{Multimodal LLMs} Currently, MLLMs have achieved breakthrough progress in architectural paradigms and general capabilities~\citep{wang2026worldtravel, wang-etal-2025-mucar}. Closed-source models like GPT-5 series~\citep{openai2025gpt5system} and Gemini 2.5 Pro~\citep{comanici2025gemini} have demonstrated remarkable capabilities, while in the open-source domain, the LLaVA series~\citep{liu2023visual} established a mainstream paradigm by aligning vision encoders with LLMs via a projection layer. Subsequent works, such as LLaVA-OneVision-1.5~\citep{an2025llava} and Qwen3-VL~\citep{Qwen3-VL}, have further optimized training strategies. However, most open-source MLLMs heavily rely on large-scale text-image pair datasets for instruction tuning. Consequently, although the language backbones of these MLLMs possess inherent multilingual capabilities, the models often suffer from performance degradation or hallucinations in non-English tasks due to the scarcity of visually aligned multilingual data.

    \paragraph{Multilingual MLLMs}
    Although recent research tries to improve MLLMs' multilingual language ability by expanding data scale, the lack of high-quality image-text data and high training costs pose a significant challenge~\citep{chen2023mclip, ruan2025layalign}.
    For instance, \citet{yue2024pangea} introduces Pangea, a series of fully open-source multilingual multimodal models. By releasing the 39-language PangeaInstruct dataset, they demonstrate that broader language coverage is key to boosting cross-cultural capabilities in MLLMs. Aya Vision~\citep{dash2025aya} utilizes high-quality synthetic data to tackle the non-English instruction following gap by capitalizing on massive text generation.
    However, these resource-intensive, data-driven approaches are hindered by the scarcity of high-quality non-English image-text pairs.
    In contrast, we address these challenges by utilizing pure text data, thereby significantly improving the multilingual performance of MLLMs.

    \paragraph{Model Merging for MLLM}
    Model merging offers a cost-efficient way to expand model capabilities without increasing inference overhead~\citep{yang2024model, zhang-etal-2025-merge, du2025adamms}. However, its potential for multilingual enhancement in MLLMs remains underexplored.
    At present, although several studies have explored enhancing MLLMs with mathematical~\citep{chenbring} or coding~\citep{jiang2025viscodex} capabilities via model merging, there is limited research on its application to multilingual adaptation. To address this gap, we propose a method that combines fine-tuning with model merging, enabling the efficient construction of multilingual MLLMs without relying on large-scale image–text datasets.

\section{Methodology}

\subsection{Vision-Free Adaptation (VFA)}

Directly fine-tuning an MLLM on multilingual text often induces modal interference, as language-only updates can disturb the vision--language alignment learned during multimodal training and weaken cross-lingual multimodal integration. As a result, improvements in multilingual text understanding do not reliably translate into stronger multilingual multimodal performance. To address this issue, VFA decouples multilingual learning from visual grounding through a task-vector formulation. Under this view, fine-tuning is treated as a task-specific parameter update that can be composed with the vision-aligned parameters of an existing MLLM.

Formally, VFA first fine-tunes the base LLM initialization $\theta_{\text{base}}$ on multilingual text data $\mathcal{D}_{\text{multi}}$ to obtain $\theta_{\text{FT}}$:
\begin{equation}
    \theta_{\text{FT}} \leftarrow \text{Tune}(\theta_{\text{base}}, \mathcal{D}_{\text{multi}}),
\end{equation}
where $\text{Tune}$ denotes either full-parameter fine-tuning or parameter-efficient adaptation~\cite{wang-etal-2025-milora}. The parameter difference $(\theta_{\text{FT}} - \theta_{\text{base}})$ is interpreted as a multilingual task vector and is injected into the MLLM backbone $\theta_{\text{MLLM}}$ via a scaling factor $\alpha$:
\begin{equation}
    \theta_{\text{merged}} = \theta_{\text{MLLM}} + \alpha (\theta_{\text{FT}} - \theta_{\text{base}}),
\end{equation}
where $\theta_{\text{merged}}$ is the resulting model parameterized for downstream tasks.

In comparison, conventional multimodal adaptation directly updates the MLLM parameters $\theta_{\text{MLLM}}$ using image--text supervision data $\mathcal{D}_{\text{visual}}$:
\begin{equation}
    \theta_{\text{tuned}} = \text{Tune}(\theta_{\text{MLLM}}, \mathcal{D}_{\text{visual}}),
\end{equation}
where $\theta_{\text{tuned}}$ represents the parameters of the conventionally fine-tuned model.

As illustrated in Figure~\ref{fig:vfa_framework}, VFA enhances multilingual capability by leveraging large-scale text-only data, while preserving visual grounding by keeping the vision encoder and projection modules frozen. Multilingual knowledge is acquired independently of visual supervision and subsequently composed with the vision-aligned MLLM, thereby mitigating interference between linguistic and visual adaptation.
This design substantially reduces reliance on scarce multilingual image--text supervision, offering a more resource-efficient alternative to joint multimodal fine-tuning. Moreover, when multiple MLLMs share the same underlying LLM, the multilingual task vector can be \textit{trained once and reused} across models via lightweight merging, whereas end-to-end multimodal adaptation must be performed separately for each MLLM, incurring significantly higher training costs.

\subsection{Merging Methods}
We consider three representative merging operators for composing the multilingual update with the vision-aligned MLLM, each offering a distinct trade-off between multilingual capability injection and preservation of vision--language alignment.

\paragraph{Weight Averaging (WA).}
Weight averaging~\citep{wortsman2022model} linearly interpolates between the pre-trained backbone and the fine-tuned parameters:
\textcolor{black}{
\begin{equation}
    \begin{aligned}
        & \Delta_{\text{FT}}  = \theta_{\text{FT}} - \theta_{\text{base}}, \\
        & \Delta_{\text{MLLM}} = \theta_{\text{MLLM}} - \theta_{\text{base}}, \\
        & \theta_{\text{merged}} = \theta_{\text{base}} + \alpha \cdot \Delta_{\text{MLLM}} + (1 - \alpha) \cdot \Delta_{\text{FT}},
    \end{aligned}
\end{equation}
where $\theta_{\text{FT}}$ and $\theta_{\text{base}}$ denote the fine-tuned LLM and the original LLM backbone within the VLM, respectively. The mixing coefficient $\alpha \in [0,1]$ balances the trade-off between general multimodal capabilities and task-specific expertise.
}

\paragraph{Task Arithmetic (TA).}
Task arithmetic~\citep{ilharcoediting} constructs a task vector $\tau$ that captures the learned update and adds it to the backbone:
\begin{equation}
    \theta_{\text{merged}} = \theta_{\text{base}} + \alpha \cdot (\Delta_\text{FT}+\Delta_\text{MLLM}),
\end{equation}
where $\alpha$ is a scaling factor used to control the strength of the injected capabilities.

\paragraph{TIES-Merging.}
To mitigate parameter interference, TIES~\citep{yadav2023ties} applies Trimming, Electing, and Sign-merging on the task vector. The merged parameters are derived as:
\textcolor{black}{
\begin{equation}
    \theta_{\text{merged}} = \theta_{\text{base}} + \alpha \cdot \text{TIES}(\Delta_\text{FT}+\Delta_\text{MLLM}),
\end{equation}
where the function $\text{TIES}(\cdot)$ represents the sequential process of: (i) Trimming to retain only the top-$k\%$ parameters by magnitude, (ii) Electing to determine the dominant sign of each parameter, and (iii) Sign-merging to aggregate values that align with the elected sign. The scaling factor $\alpha$ controls the intensity of the combined multimodal and task-specific updates.}

\subsection{Merged MLLM}
\label{sec:merged_mllm}
To ensure consistent token representations and avoid disrupting vision--language alignment during model merging, we follow the standard practice \citep{chenbring} and exclude the visual input and output embeddings from the merging process. The resulting VFA model is composed by integrating the merged language backbone with frozen visual modules:
\begin{equation}
    \theta_{\text{VFA}} = \bigl\{
    \smash{\underbrace{\theta_{\text{v-enc}}, \theta_{\text{proj}}}_{\scriptscriptstyle\text{Visual Backbone}}},\;
    \smash{\underbrace{\theta_{\text{emb}}^{\text{m}}, \theta_{\text{trans}}^{\text{m}}, \theta_{\text{head}}^{\text{m}}}_{\scriptscriptstyle\text{Merged Language Backbone}}}
    \bigr\},
\end{equation}
\vspace{0.1em}

\noindent where the superscript ${\text{m}}$ denotes merged parameters. Specifically, $\theta_{\text{v-enc}}$ and $\theta_{\text{proj}}$ represent the visual encoder and vision language projector inherited from the original MLLM, respectively. For the language backbone, $\theta_{\text{emb}}^{\text{m}}$, $\theta_{\text{trans}}^{\text{m}}$, and $\theta_{\text{head}}^{\text{m}}$ correspond to the merged token embeddings, transformer blocks, and output head. By keeping the visual stream fixed, this design retains the model’s vision--language grounding while enabling the seamless integration of multilingual knowledge without additional training.
\textcolor{black}{Importantly, VFA produces a single unified model with no additional inference latency or memory overhead compared to the original MLLM.}

\input{tables/tab_01_eval_benchmarks}

\section{Experiments}

\input{tables/tab_02_multilingual_multimodal_results}

\subsection{Experimental Setup}
\paragraph{Multilingual Adaptation.}
\textbf{(1) Data.} For multilingual adaptation, we fine-tune the base LLMs on a 100K-example subset of the Multilingual-SFT dataset,\footnote{https://huggingface.co/datasets/agentlans/multilingual-sft} which combines xP3mt~\citep{muennighoff2022crosslingual}, Bactrian-X~\citep{li2023bactrianx}, the text subset of Aya Vision~\citep{singh2024aya}, and LLM-generated instruction data.
\textbf{(2) Models.} We evaluate VFA on five MLLMs spanning multiple model families and scales, including Qwen2.5~\citep{qwen2.5}, Qwen3~\citep{qwen3technicalreport}, Llama3~\citep{llama3modelcard}, and Llama3.1~\citep{meta2024llama31} series (4B--8B). \textbf{(3) Training.} All experiments are run on 4$\times$A100 (80GB) GPUs using LlamaFactory~\citep{zheng2024llamafactory}. Detailed dataset descriptions, models, and hyperparameters are provided in Appendix~\ref{sec:appendix-training-settings}.

\paragraph{Multimodal Model Merging.}
To incorporate multilingual capability into MLLMs, we compare three merging methods (WA, TA, and TIES). For each model pair, we conduct experiments with mixing coefficients $\alpha \in \{0.5, 0.7, 0.9, 1.0\}$ and utilize CVQA~\citep{mogrovejo2024cvqa} as a validation set to select the best combination.

\paragraph{Evaluation.}
For a comprehensive assessment, we evaluate models along three axes: (i) multilingual multimodal capability, (ii) general multimodal capability, and (iii) text-only multilingual capability. Table~\ref{tab:eval_benchmarks} summarizes the benchmark suite for each axis. We utilize OpenCompass~\citep{2023opencompass} framework for text-only benchmarks and lmms-eval~\citep{zhang2024lmmsevalrealitycheckevaluation} framework for multimodal benchmarks, using vLLM for inference. We set the maximum sequence length to 2048 and the batch size to 512.
Additional details are provided in Appendix~\ref{sec:appendix-llm-benchmarks}.

\input{tables/tab_03_general_multimodal_results}

\subsection{Multilingual Multimodal Results}

As shown in Table~\ref{tab:merged_multilingual_results}, we compare VFA with the original MLLM (\textit{Base}) and direct text-only fine-tuning (\textit{Direct Text SFT}).
Across all evaluated models, VFA consistently improves multilingual multimodal performance under text-only supervision, leading to higher overall averages without degrading existing capabilities. The most pronounced gains are observed on culturally grounded visual reasoning tasks (MaRVL), with improvements of +12.33 on Qwen2.5-VL-7B and +36.17 on Idefics3-8B. This pattern indicates that VFA is particularly effective for language-intensive cross-cultural grounding, where accurate linguistic interpretation plays a central role in visual reasoning. Similar trends hold across model families and scales, yielding average gains of +2.39 (Qwen2.5-VL-7B), +10.64 (Idefics3-8B), and +6.35 (LLaVA-Next-8B), as well as substantial improvements on Qwen3-based LLaVA-OV models (+2.99 for 8B and +0.57 for 4B), with notable gains on xGQA.
Per-language breakdowns for all benchmarks and qualitative examples are provided in Appendix~\ref{sec:appendix-mllm-benchmarks} and~\ref{sec:qualitative analysis}, respectively.


In contrast, direct text-only fine-tuning frequently leads to degraded multilingual multimodal performance, reflecting interference between linguistic adaptation and pre-existing vision--language alignment. For Qwen2.5-VL-7B, Direct Text SFT reduces the overall average by 11.92 and causes a near-complete collapse on MaRVL (from 53.50 to 0.00), despite remaining competitive on several other benchmarks. A similar trend is observed for LLaVA-Next-8B: although Direct Text SFT improves MaXM and XM100, it substantially degrades performance on MaRVL (-39.83), resulting in an overall average drop of 3.40. Taken together, these results align with the motivation of VFA: composing a multilingual task vector with a vision-aligned MLLM enables broad multilingual gains while mitigating the alignment degradation that can arise when the MLLM backbone is directly adapted using text-only supervision. 

\input{tables/tab_04_multilingual_text_task_results}

\begin{figure*}[htbp]
    \centering
    \includegraphics[width=1.0\linewidth]{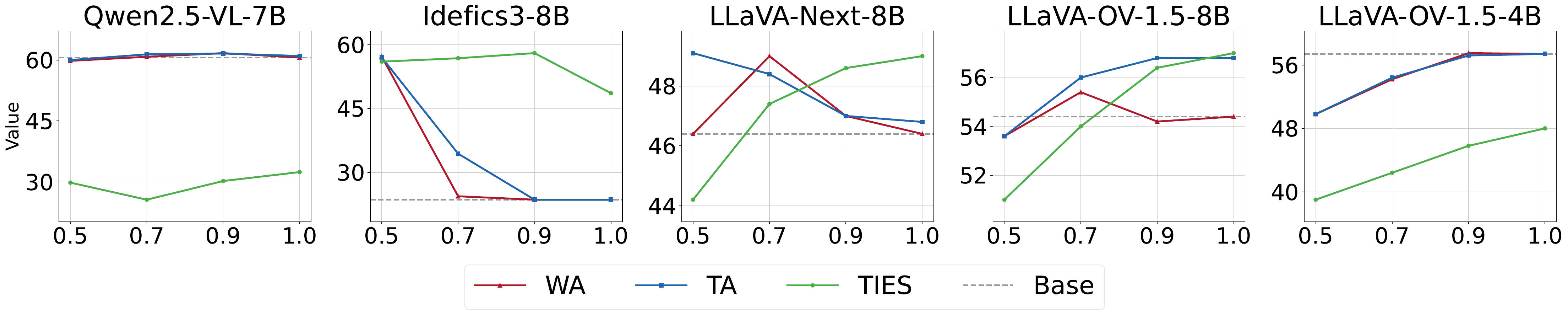}
    \vspace{-15pt}
    \caption{\textbf{Effect of Merging Operators and Mixing Coefficient.} Validation performance on CVQA under different merging operators (WA, TA, and TIES) and mixing coefficients \(\alpha\).}
    \label{fig:merge_strategies_cvqa}
    \vspace{-10pt}
\end{figure*}

\subsection{General Multimodal Results}
As shown in Table~\ref{tab:general_multimodal_comparison_updated}, VFA maintains or improves \textit{general-purpose vision--language performance}. Overall, we observe no systematic degradation: VFA remains close to neutral on average for most models and can be beneficial in some cases, such as +1.36 on Idefics3-8B and +1.00 on Qwen2.5-VL-7B, while staying nearly unchanged on LLaVA-OV-1.5-8B (--0.19). This pattern indicates that introducing multilingual capability through VFA does not inherently interfere with the visual grounding learned during multimodal training.

The observed gains are not uniform across benchmarks. Improvements are primarily concentrated on reasoning-centric evaluations such as MMMU and MathVista, while MMBench and OCRBench exhibit only modest, backbone-dependent variations. For instance, the average improvement on Idefics3-8B is driven largely by a marked gain on MathVista (+7.80), despite minor declines on OCRBench and MMBench. This pattern suggests that enhancing the language backbone predominantly benefits language-intensive multimodal reasoning, including problem interpretation and multi-step inference, whereas perception-heavy skills such as OCR are less directly affected. Incorporating additional multimodal supervision may therefore be necessary to further improve performance on perception-focused benchmarks.

\subsection{Text-only Multilingual Results}
\label{sec:text_only_results}

\input{tables/tab_05_pangea_comparison}

Table~\ref{tab:text_only_ft_comparison_MLLM} presents text-only multilingual performance after injecting multilingual knowledge into MLLMs via VFA. Across both Qwen and Llama model families, VFA preserves or improves performance on the vast majority of benchmarks. In particular, it brings modest gains to Qwen-based models and delivers substantial improvements for LLaVA-Next-8B ($+6.27$ on average). The main outlier is Idefics3-8B, which achieves strong gains on reasoning and knowledge-intensive tasks but exhibits a pronounced decline on FLORES ($-13.94$) and a smaller drop on XNLI ($-3.08$).

Inspecting outputs on the FLORES translation task reveals that this regression is primarily an artifact of generation behavior rather than a fundamental loss of translation ability. The merged Idefics3 model typically translates the first sentence correctly but fails to emit the EOS token, over-generating hallucinated text until the length limit. Since FLORES uses single-sentence references, BLEU heavily penalizes this via precision dilution, drastically suppressing the score despite the accurate initial translation. This issue is most pronounced in languages with concise references (e.g., Chinese) and is absent in LLaVA-Next-8B, suggesting that the EOS-control mechanisms of underlying instruction-tuned LLMs react differently to weight merging across MLLM families. Representative examples of this EOS-failure pattern are in Appendix~\ref{app:flores_generation}.



\input{tables/tab_06_visual_coding_results}

\section{Analysis}

\subsection{Merging Strategies}

We analyze how merging choices affect the merged MLLM. Figure~\ref{fig:merge_strategies_cvqa} compares the validation performance (CVQA) of three operators (WA, TA, and TIES) across different mixing coefficients $\alpha$. The results indicate that the performance trends are highly dependent on architecture. Specifically, for Qwen2.5-VL-7B and LLaVA-OV-1.5-4B, WA and TA perform similarly and significantly outperform TIES. In contrast, for Idefics3-8B, TIES demonstrates stronger overall performance, while WA and TA experience substantial degradation at higher coefficients. For LLaVA-Next-8B and LLaVA-OV-1.5-8B, the optimal operator varies with $\alpha$, showing no single dominant method across the range. Consequently, rather than applying a universal setting, our final configuration for each backbone is determined by selecting the specific combination of operator and coefficient $\alpha$ that yields the highest absolute validation score. Detailed parameter choices and recommended defaults are summarized in Appendix~\ref{sec:recommended-defaults}.

\subsection{Comparison with Multimodal Fine-tuning}

We compare VFA to conventional multimodal training that relies on large-scale paired image--text supervision. As shown in Table~\ref{tab:main_multimodal_results}, Pangea-7B is trained with 6M multimodal samples on Qwen2-7B-Instruct and achieves 50.34 on average, whereas VFA starts from Qwen2-VL-7B and uses only 100K text-only samples (about 2\% of Pangea's training size) to reach 47.84, leaving a 2.50-point gap. From an efficiency-frontier perspective, VFA shifts the performance vs. data-cost trade-off leftward by replacing expensive multilingual image--text curation with rich multilingual text. The remaining gap is plausibly attributable to capabilities that benefit from paired supervision, such as stronger visual grounding and broader multilingual vision--language alignment coverage; closing it may require higher-quality multilingual text, stronger language backbones, or lightweight hybrid objectives that reintroduce limited multimodal signals.

\subsection{Beyond Multilingual: VFA on Visual Coding Tasks}
While we primarily focus on multilingual adaptation, VFA is inherently a versatile vision-free framework. To illustrate its broader applicability, we present an additional case study applying the vision-free injection paradigm to multimodal visual coding. As shown in Table~\ref{tab:VFA_summary_large_font}, VFA achieves consistent performance improvements across all three evaluated backbones, yielding the most significant gains on InternVL3-8B (+4.59) and Idefics3-8B (+6.62). These findings underscore VFA's potential as a lightweight, reusable mechanism for expanding MLLM capabilities beyond multilingual contexts.



\section{Conclusion}
In this paper, we introduce VFA, an efficient framework designed to enhance the multilingual capabilities of MLLMs without relying on additional text-image paired datasets. Additionally, VFA decouples linguistic knowledge injection from visual representation learning, effectively circumventing the catastrophic forgetting of visual alignment that often plagues direct text fine-tuning methods. Extensive evaluations across five diverse architectures and six multilingual multimodal benchmarks show that VFA not only preserves generic multimodal robustness but also surpasses models trained on massive multimodal datasets in some tasks. Overall, this work provides a \textcolor{black}{practical and} efficient pathway to broaden the linguistic capabilities of MLLMs.


\section*{Limitations}

The multilingual data used for adaptation does not yet incorporate explicit quality filtering, and its scale is intentionally kept modest to highlight the efficiency of VFA. In addition, as a text-only approach, VFA primarily enhances language-intensive reasoning, while perception-heavy tasks such as OCR may still benefit from supplementary image–text supervision. Our empirical evaluation mainly focuses on models in the 4B--8B parameter range. Exploring larger model and data scales, adopting stricter data curation, and developing more advanced merging algorithms to more consistently mitigate this interference remain important directions for future work.


\section*{Acknowledgements}

This project was supported by National Key R\&D Program of China (No. 2025YFB4007600), National Natural Science Foundation of China (No. 62306132), Guangdong Basic and Applied Basic Research Foundation (No. 2025A1515011564), Natural Science Foundation of Shanghai (No. 25ZR1402136). We thank the anonymous reviewers for their insightful feedback on this work.

\bibliography{custom}

\appendix

\textcolor{black}{
\begin{table*}[h!]
    \centering
    \small
    \setlength{\tabcolsep}{3pt}
    \begin{tabular}{@{} llcrr @{}}
        \toprule
        \textbf{Category} & \textbf{Source} & \textbf{\#Sub} & \textbf{Count} & \textbf{Prop.} \\
        \midrule
        Primary large-scale & Aya, Bactrian-X, xP3mt, Tagengo & 4 & 92,907 & 92.91\% \\
        Multi. Alpaca & FreedomIntel. (Alpaca-GPT4) & 11 & 4,965 & 4.97\% \\
        Multi. Evol & FreedomIntel. (Evol-Instruct) & 11 & 2,128 & 2.13\% \\
        \midrule
        \textbf{Total} & & \textbf{26} & \textbf{100,000} & \textbf{100\%} \\
        \bottomrule
    \end{tabular}
    \caption{Composition of the 100K multilingual text subset.}
    \label{tab:data_composition}
\end{table*}
}

\input{tables/tab_07_language_codes}
\input{tables/tab_08_training_efficiency}
\input{tables/tab_09_p_mmeval_benchmark}

\section{Training Settings}
\label{sec:appendix-training-settings}
\subsection{Datasets}
\paragraph{Data Construction.}
\textcolor{black}{The 100K multilingual text subset is constructed deterministically from the \texttt{train} split of the HuggingFace dataset \texttt{agentlans/multilingual-sft} (configuration: \texttt{clustered\_k100000}), traversed in its default order without random sampling. We apply the following fixed filtering rules: (1) Field validation: samples are excluded if the input or source fields are not valid strings; (2) Multimodal token filtering: instances containing <audio>, <image>, or <video> tags are removed to ensure pure text-only supervision. The resulting dataset is summarized in Table~\ref{tab:data_composition}, and Table~\ref{tab:language_codes} further lists the ISO 639-1 language codes represented.}

\paragraph{xP3mt.} xP3mt~\citep{muennighoff2022crosslingual} is a large-scale multilingual instruction tuning dataset. Derived from the xP3 dataset, xP3mt aims to address the scarcity of non-English prompts by utilizing machine translation to translate English prompts into 20 diverse languages. The dataset includes 46 languages and 13 training tasks, including QA, Summarization, and Program Synthesis.
\paragraph{Bactrian-X.} Bactrian-X~\citep{li2023bactrianx} is a comprehensive multilingual instruction dataset designed to democratize instruction following capabilities across 52 languages. It includes approximately 3.4 million instruction-response pairs. The dataset was constructed by translating English instructions from Alpaca and Dolly-15k into 51 target languages, followed by feeding these translated prompts into gpt-3.5-turbo to generate authentic, native-like responses.
\paragraph{Aya Dataset.} The Aya Dataset~\citep{singh2024aya} is a large-scale human-curated multilingual instruction tuning dataset, comprising approximately 204K instruction-response pairs covering 65 languages. Its human-centric curation helps mitigate the biases and artifacts commonly found in machine-translated corpora, particularly for underrepresented low-resource languages.

\input{tables/tab_10_multimodal_benchmarks}

\paragraph{Tagengo-GPT4.} Tagengo-GPT4~\citep{devine2024tagengo} is a high-quality multilingual instruction tuning dataset developed by Lightblue, explicitly designed to bridge the gap between English-centric LLMs and global language accessibility. The dataset includes approximately 76,000 diverse prompt-response pairs covering 74 languages.

\paragraph{Multilingual-SIFT.} Multilingual-SIFT~\citep{Chen_MultilingualSIFT_Multilingual_Supervised_2023} is a comprehensive dataset collection specifically constructed to enhance the multilingual instruction-following capabilities of LLMs. It is derived from high-quality English instruction corpora, including Alpaca-GPT4, Evol-Instruct, and ShareGPT. To ensure linguistic diversity, the authors employed GPT-3.5 Turbo to translate these datasets into multiple target languages.

\subsection{Models}
\label{sec: appendix-models-detail}
Table~\ref{tab:model_specifications} presents the MLLM models used in this paper and their corresponding LLMs.
\input{tables/tab_11_model_backbones}

\subsection{Hyperparameters}
For the fine-tuning experiments, we set the total batch size to 128, with a learning rate of 1e-5, and train for 1 epoch.

\textcolor{black}{
\subsection{Training Efficiency}
Table~\ref{tab:efficiency_comparison} compares the training efficiency between standard multimodal SFT and VFA. By relying solely on text-only data and avoiding visual token processing, VFA reduces training time, GPU memory consumption, and data requirements, making it a practical and scalable alternative to conventional multilingual multimodal fine-tuning.
}

\input{tables/tab_12_selected_merge_configs}

\begin{figure*}[htbp]
    \centering
    \includegraphics[width=1.0\linewidth]{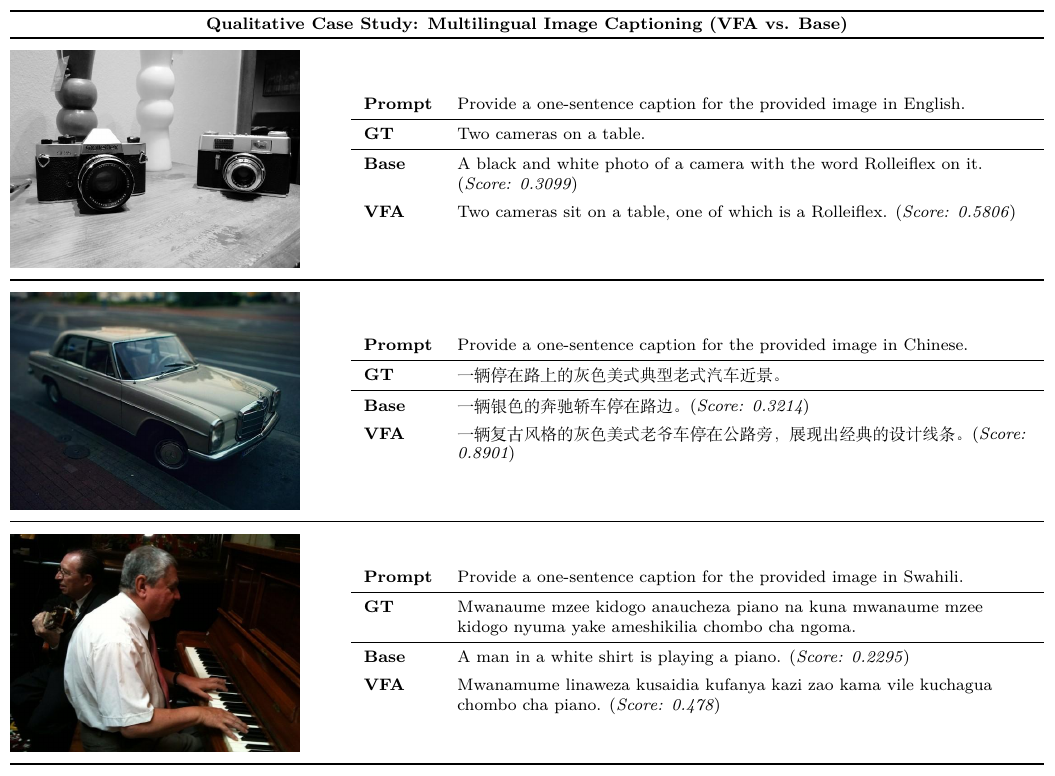}
    \caption{Qualitative comparison between VFA and base MLLM on XM100.}
    \label{fig:qualitative_vfa_case}
\end{figure*}

\section{Evaluation}
\label{sec:appendix-llm-benchmarks}
We compare VFA-enhanced MLLMs with baselines on the following benchmarks:

\paragraph{TyDiQA.} TyDiQA~\citep{clark2020tydi} is a benchmark for evaluating QA systems across 11 typologically diverse languages, comprising approximately 200K human-annotated QA pairs.

\paragraph{P-MMEval.} P-MMEval~\citep{zhang2025p} is a large-scale parallel multilingual multitask benchmark specifically designed to enable consistent and fair evaluation of LLMs across diverse linguistic landscapes. It provides strictly parallelized evaluation samples across 10 typologically distinct languages. As shown in Table~\ref{tab:P-MMEval-benchmarks}, the benchmark integrates fundamental NLP tasks with capability-specialized challenges, serving as a rigorous testbed for assessing a model's cross-lingual transferability and core reasoning proficiency without the noise of dataset inconsistency.

\begin{figure*}[htbp]
\centering
\includegraphics[width=\linewidth]{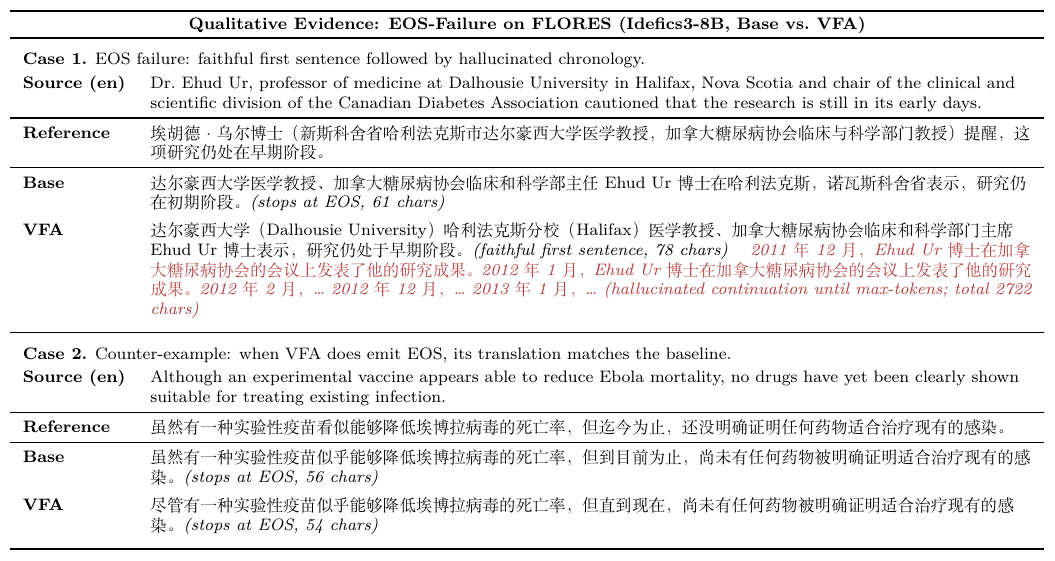}
\caption{\textbf{Comparison of Outputs from the Idefics3-8B Base MLLM and VFA on FLORES Chinese.} \textbf{Case 1} illustrates the EOS failure driving the BLEU regression: VFA translates the first sentence accurately but hallucinates thereafter until the length limit. \textbf{Case 2} shows proper EOS emission, yielding a translation comparable to the base MLLM. Hallucinations are in \textcolor{DarkRed}{red}.}
\label{fig:flores_eos_failure}
\end{figure*}

\paragraph{PangeaBench.} PangeaBench~\citep{yue2024pangea} is a comprehensive evaluation designed to assess the multilingual and multicultural capabilities of MLLMs. Including 47 languages across 14 diverse datasets, PangeaBench extends beyond standard translation-based metrics by incorporating culturally specific tasks. As shown in Table~\ref{tab:multimodal_benchmark}, the benchmark spans five primary categories, ranging from multimodal chat to multi-subject reasoning.

\section{PangeaBench Results by Language}
\label{sec:appendix-mllm-benchmarks}
We present the performance of the VFA-enhanced MLLM and the base MLLM on the MaXM, xGQA, xMMMU, MaRVL, M3Exam, and XM100 benchmarks in Tables~\ref{tab:maxm_results}--\ref{tab:breakdown_xm100}.

\section{Qualitative Analysis}
\label{sec:qualitative analysis}
Furthermore, to offer a more intuitive perspective, we present qualitative comparisons between VFA-equipped MLLMs and standard base models, as illustrated in Figure~\ref{fig:qualitative_vfa_case}. Through various multilingual visual question answering tasks, these examples vividly demonstrate how VFA practically enhances the precision and overall quality of non-English generations. Notably, these gains avoid the typical vision-language alignment tax.

\input{tables/tab_13_maxm_results}
\input{tables/tab_14_xgqa_results}
\input{tables/tab_15_xmmmu_results}
\input{tables/tab_16_marvl_results}
\input{tables/tab_17_m3exam_results}
\input{tables/tab_18_xm100_results}

\section{Generation Behavior on FLORES}
\label{app:flores_generation}

As discussed in Section \ref{sec:text_only_results}, Figure \ref{fig:flores_eos_failure} provides raw output examples illustrating the EOS-failure pattern in Idefics3-8B. While both the base model and VFA model successfully generate accurate first-sentence translations, the VFA model frequently fails to emit the EOS token. Consequently, it over-generates hallucinated text until reaching the length limit (as seen in Example 1). This qualitative evidence confirms that the underlying translation capability is preserved, and the observed BLEU regression is primarily a generation-length artifact.

\section{Recommended Merge Defaults}
\label{sec:recommended-defaults}

As shown in Table~\ref{tab:selected_merge_configs}, we establish practical guidelines for model merging based on our empirical findings.

\section{Monolingual Adaptation}
\label{sec:monolingual adaptation}
To demonstrate the versatility of our method, beyond the multilingual setting, we also evaluate VFA under monolingual adaptation by individually fine-tuning on Hindi and Romanian—two languages not included in our main multilingual suite. Results are reported in Table~\ref{tab:multilingual_hi_ro_combined}, showing that VFA remains effective for targeted, language-specific adaptation.

\input{tables/tab_19_hi_ro_results}


\end{document}

%% file: tables/tab_01_eval_benchmarks.tex
\begin{table}[t]
\centering

    \setlength{\tabcolsep}{4pt} 
    \setlength{\aboverulesep}{0pt}
    \setlength{\belowrulesep}{0pt}
    \renewcommand{\arraystretch}{1.2}

\footnotesize
\renewcommand{\arraystretch}{1.15}
\begin{tabularx}{\linewidth}{@{}>{\raggedright\arraybackslash}p{0.30\linewidth} >{\raggedright\arraybackslash}X@{}}
\toprule
\textbf{Dimension} & \textbf{Benchmarks} \\
\midrule
\textsc{Multilingual multimodal} & MaXM, xGQA, xMMMU, XM100, MaRVL, M3Exam \\
\addlinespace[2pt]
\textsc{General multimodal} & OCRBench, MMBench, MMMU, MathVista \\
\addlinespace[2pt]
\textsc{Text-only multilingual} & TyDiQA, MMMLU, XNLI, HellaSwag, MLogiQA, M-IFEval, FLORES \\
\bottomrule
\end{tabularx}
\caption{Benchmark suite grouped by evaluation axis.}
\label{tab:eval_benchmarks}
\vspace{-0.4cm}
\end{table}

%% file: tables/tab_02_multilingual_multimodal_results.tex
\begin{table*}[htbp]
\centering
\setlength{\tabcolsep}{4mm}
\resizebox{1.0\linewidth}{!}{
    \large
    \begin{tabular}{l|l|cccccc|c}
    \toprule
    \textbf{Model} & \textbf{Method} & \textbf{MaXM} & \textbf{xGQA} & \textbf{xMMMU} & \textbf{XM100} & \textbf{MaRVL} & \textbf{M3Exam} & \textbf{Avg.} \\ \hline

    \multirow{5}{*}{\shortstack[l]{Qwen2.5-VL-7B \\ \footnotesize \color{gray} \firstarrow Qwen2.5-7B-base}}
     & \gcell Base & \gcell 50.37 & \gcell 47.81 & \gcell 48.34 & \gcell 15.74 & \gcell 53.50 & \gcell 61.97 & \gcell 46.29 \\
     & Direct Text SFT & 49.68 & 31.20 & 49.69 & 13.80 & 0.00 & 61.86 & 34.37 \\
     & \textit{\small $\Delta$ Gains} & \rbf{--0.69} & \rbf{--16.61} & \gbf{+1.35} & \rbf{--1.94} & \rbf{--53.50} & \rbf{--0.11} & \rbf{--11.92} \\
     & \bcell VFA (ours) & \bcell 51.69 & \bcell 48.45 & \bcell 48.15 & \bcell 15.43 & \bcell 65.83 & \bcell 62.55 & \bcell 48.68 \\
     & \textit{\small $\Delta$ Gains} & \gbf{+1.32} & \gbf{+0.64} & \rbf{--0.19} & \rbf{--0.31} & \gbf{+12.33} & \gbf{+0.58} & \gbf{+2.39} \\ \hline

    \multirow{5}{*}{\shortstack[l]{Idefics3-8B \\ \footnotesize \color{gray} \firstarrow Llama3.1-8B-inst}}
     & \gcell Base & \gcell 46.40 & \gcell 48.73 & \gcell 42.83 & \gcell 13.80 & \gcell 26.50 & \gcell 27.93 & \gcell 34.37 \\
     & Direct Text SFT & 46.56 & 44.95 & 36.45 & 14.06 & 38.00 & 44.79 & 37.47 \\
     & \textit{\small $\Delta$ Gains} & \gbf{+0.16} & \rbf{--3.78} & \rbf{--6.38} & \gbf{+0.26} & \gbf{+11.50} & \gbf{+16.86} & \gbf{+3.10} \\
     & \bcell VFA (ours) & \bcell 50.11 & \bcell 48.10 & \bcell 43.27 & \bcell 16.20 & \bcell 62.67 & \bcell 49.73 & \bcell 45.01 \\
     & \textit{\small $\Delta$ Gains} & \gbf{+3.71} & \rbf{--0.63} & \gbf{+0.44} & \gbf{+2.40} & \gbf{+36.17} & \gbf{+21.80} & \gbf{+10.64} \\ \hline

    \multirow{5}{*}{\shortstack[l]{LLaVA-Next-8B \\ \footnotesize \color{gray} \firstarrow Llama3-8B-inst}}
     & \gcell Base & \gcell 30.26 & \gcell 43.60 & \gcell 37.73 & \gcell 1.63 & \gcell 46.33 & \gcell 45.48 & \gcell 34.17 \\
     & Direct Text SFT & 42.43 & 44.01 & 35.50 & 13.92 & 6.50 & 42.24 & 30.77 \\
     & \textit{\small $\Delta$ Gains} & \gbf{+12.17} & \gbf{+0.41} & \rbf{--2.23} & \gbf{+12.29} & \rbf{--39.83} & \rbf{--3.24} & \rbf{--3.40} \\
     & \bcell VFA (ours) & \bcell 42.80 & \bcell 45.39  & \bcell 38.62 & \bcell 14.43 & \bcell 55.33 & \bcell 46.57 & \bcell 40.52 \\
     & \textit{\small $\Delta$ Gains} & \gbf{+12.54} & \gbf{+1.79} & \gbf{+0.89} & \gbf{+12.80} & \gbf{+9.00} & \gbf{+1.09} & \gbf{+6.35} \\ \hline

    \multirow{5}{*}{\shortstack[l]{LLaVA-OV-1.5-8B \\ \footnotesize \color{gray} \firstarrow Qwen3-8B-base}}
    & \gcell Base & \gcell 53.97 & \gcell 29.85 & \gcell 54.71 & \gcell 16.51 & \gcell 62.17 & \gcell 63.86 & \gcell 46.85 \\
    & Direct Text SFT & 52.92 & 28.73 & 55.80 & 15.07 & 61.83 & 64.08 & 46.41 \\
    & \textit{\small $\Delta$ Gains} & \rbf{--1.05} & \rbf{--1.12} & \gbf{+1.09} & \rbf{--1.44} & \rbf{--0.34} & \gbf{+0.22} & \rbf{--0.44} \\
     & \bcell VFA (ours) & \bcell 56.14 & \bcell 43.97  & \bcell 53.67  & \bcell 16.02  & \bcell 65.50  & \bcell 63.75 & \bcell 49.84 \\
     & \textit{\small $\Delta$ Gains} & \gbf{+2.17} & \gbf{+14.12} & \rbf{--1.04} & \rbf{--0.49} & \gbf{+3.33} & \rbf{--0.11} & \gbf{+2.99} \\ \hline

    \multirow{5}{*}{\shortstack[l]{LLaVA-OV-1.5-4B \\ \footnotesize \color{gray} \firstarrow Qwen3-4B-base}}
     & \gcell Base & \gcell 49.52 & \gcell 36.49 & \gcell 53.44 & \gcell 14.86 & \gcell 60.50 & \gcell 59.79 & \gcell 45.77 \\
     & Direct Text SFT & 49.47 & 32.17 & 53.05 & 14.32 & 60.33 & 59.75 & 44.85  \\
    & \textit{\small $\Delta$ Gains} & \rbf{--0.05} & \rbf{--4.32} & \rbf{--0.39} & \rbf{--0.54} & \rbf{--0.17} & \rbf{--0.04} & \rbf{--0.92} \\
     & \bcell VFA (ours) & \bcell 51.64  & \bcell 37.71 & \bcell 53.42 & \bcell 14.34  & \bcell 60.67 & \bcell 60.26 & \bcell 46.34 \\
     & \textit{\small $\Delta$ Gains} & \gbf{+2.12} & \gbf{+1.22} & \rbf{--0.02} & \rbf{--0.52} & \gbf{+0.17} & \gbf{+0.47} & \gbf{+0.57} \\
    \bottomrule
    \end{tabular}
}
\caption{\textbf{Multilingual Multimodal Results of VFA across Various MLLMs.} \gbf{Green} and \rbf{Red} values in $\Delta$ Gains rows denote relative changes compared to the base MLLM. \firstarrow indicates the backbone of the MLLM.}
\label{tab:merged_multilingual_results}

\end{table*}

%% file: tables/tab_03_general_multimodal_results.tex
\begin{table*}[t]

    \centering
    \small

    \setlength{\tabcolsep}{4pt}
    \setlength{\aboverulesep}{0pt}
    \setlength{\belowrulesep}{0pt}
    \renewcommand{\arraystretch}{1.2}

    \resizebox{0.8\linewidth}{!}{
    \begin{tabular}{l|l|cccc|c}
        \toprule
        \textbf{Model} & \textbf{Method} & \textbf{OCRBench} & \textbf{MMBench} & \textbf{MMMU} & \textbf{MathVista} & \textbf{Avg.} \\
        \midrule

        \multirow{5}{*}{\shortstack[l]{Qwen2.5-VL-7B \\ \footnotesize \color{gray} \firstarrow Qwen2.5-7B-base}}
         & \gcell Base & \gcell 14.36 & \gcell 87.80 & \gcell 51.11 & \gcell 61.90 & \gcell 53.79 \\
         & Direct Text SFT & 14.56 & 87.20 & 51.33 & 65.90 & 54.75 \\
         & \textit{\small $\Delta$ Gains} & \gbf{+0.20} & \rbf{--0.60} & \gbf{+0.22} & \gbf{+4.00} & \gbf{+0.96} \\
         & \bcell VFA & \bcell 14.47 & \bcell 87.20 & \bcell 51.78 & \bcell 65.70 & \bcell 54.79 \\
         & \textit{\small $\Delta$ Gains} & \gbf{+0.11} & \rbf{--0.60} & \gbf{+0.67} & \gbf{+3.80} & \gbf{+1.00} \\
        \midrule

        \multirow{5}{*}{\shortstack[l]{Idefics3-8B \\ \footnotesize \color{gray} \firstarrow Llama3.1-8B-inst}}
         & \gcell Base & \gcell 5.96 & \gcell 84.50 & \gcell 38.78 & \gcell 24.00 & \gcell 38.31 \\
         & Direct Text SFT & 6.90 & 83.70 & 38.11 & 33.40 & 40.53 \\
         & \textit{\small $\Delta$ Gains} & \gbf{+0.94} & \rbf{--0.80} & \rbf{--0.67} & \gbf{+9.40} & \gbf{+2.22} \\
         & \bcell VFA & \bcell 4.50 & \bcell 82.50 & \bcell 39.89 & \bcell 31.80 & \bcell 39.67 \\
         & \textit{\small $\Delta$ Gains} & \rbf{--1.46} & \rbf{--2.00} & \gbf{+1.11} & \gbf{+7.80} & \gbf{+1.36} \\
        \midrule

        \multirow{5}{*}{\shortstack[l]{LLaVA-Next-8B \\ \footnotesize \color{gray} \firstarrow Llama3-8B-inst}}
         & \gcell Base & \gcell 6.44 & \gcell 79.60 & \gcell 38.00 & \gcell 20.60 & \gcell 36.16 \\
         & Direct Text SFT & 6.13 & 78.90 & 38.67 & 24.20 & 36.98 \\
         & \textit{\small $\Delta$ Gains} & \rbf{--0.31} & \rbf{--0.70} & \gbf{+0.67} & \gbf{+3.60} & \gbf{+0.82} \\
         & \bcell VFA & \bcell 6.40 & \bcell 80.10 & \bcell 40.11 & \bcell 20.80 & \bcell 36.85 \\
         & \textit{\small $\Delta$ Gains} & \rbf{--0.04} & \gbf{+0.50} & \gbf{+2.11} & \gbf{+0.20} & \gbf{+0.69} \\
        \midrule

        \multirow{5}{*}{\shortstack[l]{LLaVA-OV-1.5-8B \\ \footnotesize \color{gray} \firstarrow Qwen3-8B-base}}
         & \gcell Base & \gcell 12.50 & \gcell 88.60 & \gcell 55.33 & \gcell 32.80 & \gcell 47.31 \\
         & Direct Text SFT & 5.26 & 98.00 & 55.00 & 38.33 & 49.15 \\
        & \textit{\small $\Delta$ Gains} & \rbf{--7.24} & \gbf{+9.40} & \rbf{--0.33} & \gbf{+5.53} & \gbf{+1.84} \\
         & \bcell VFA & \bcell 12.34 & \bcell 88.10 & \bcell 56.22 & \bcell 31.80 & \bcell 47.12 \\
         & \textit{\small $\Delta$ Gains} & \rbf{--0.16} & \rbf{--0.50} & \gbf{+0.89} & \rbf{--1.00} & \rbf{--0.19} \\
        \midrule

        \multirow{5}{*}{\shortstack[l]{LLaVA-OV-1.5-4B \\ \footnotesize \color{gray} \firstarrow Qwen3-4B-base}}
         & \gcell Base & \gcell 11.44 & \gcell 89.90 & \gcell 53.22 & \gcell 36.50 & \gcell 47.77 \\
        & Direct Text SFT & 5.14 & 99.00 & 53.00 & 32.33 & 47.37 \\
        & \textit{\small $\Delta$ Gains} & \rbf{--6.30} & \gbf{+9.10} & \rbf{--0.22} & \rbf{--4.17} & \rbf{--0.40} \\
         & \bcell VFA & \bcell 11.44 & \bcell 90.10 & \bcell 54.22 & \bcell 37.50 & \bcell 48.32 \\
         & \textit{\small $\Delta$ Gains} & \gbf{+0.00} & \gbf{+0.20} & \gbf{+1.00} & \gbf{+1.00} & \gbf{+0.55} \\
        \bottomrule
    \end{tabular}
    }
       \caption{\textbf{General Multimodal Results of VFA across Various MLLMs.} \gbf{Green} and \rbf{Red} values in $\Delta$ Gains rows denote relative changes compared to the base MLLM. \firstarrow indicates the backbone of the MLLM.}
           \label{tab:general_multimodal_comparison_updated}

    \end{table*}

%% file: tables/tab_04_multilingual_text_task_results.tex
\begin{table*}[t]
    \centering

    \setlength{\tabcolsep}{4pt} 
    \setlength{\aboverulesep}{0pt}
    \setlength{\belowrulesep}{0pt}
    \renewcommand{\arraystretch}{1.2}
    
\resizebox{1.0\linewidth}{!}{
    \large
    \begin{tabular}{l|l|ccccccc|c}
    \toprule
    \textbf{Model} & \textbf{Method} & \textbf{TyDIQA} & \textbf{MMMLU} & \textbf{XNLI} & \textbf{HellaSwag} & \textbf{MLogiQA} & \textbf{M-IFEval} & \textbf{FLORES} & \textbf{Avg.} \\ 
    \midrule

    \multirow{3}{*}{\shortstack[l]{Qwen2.5-VL-7B \\ \footnotesize \color{gray} \firstarrow Qwen2.5-7B-base}}
    & \gcell Base & \gcell 24.89 & \gcell 44.70 & \gcell 67.67 & \gcell 59.02 & \gcell 46.62 & \gcell 66.56 & \gcell 37.04 & \gcell 49.50 \\
    & \bcell VFA & \bcell 31.05  & \bcell 48.85 & \bcell 68.59 & \bcell 60.31 & \bcell 48.88 & \bcell 66.67 & \bcell 33.90 & \bcell 51.18 \\
    & \textit{\small $\Delta$ Gains} & \gbf{+6.16} & \gbf{+4.15} & \gbf{+0.92} & \gbf{+1.29} & \gbf{+2.26} & \gbf{+0.11} & \rbf{--3.14} & \gbf{+1.68} \\ 
    \midrule

    \multirow{3}{*}{\shortstack[l]{Idefics3-8B \\ \footnotesize \color{gray} \firstarrow Llama3.1-8B-inst}}
    & \gcell Base & \gcell 31.60 & \gcell 45.35 & \gcell 63.83 & \gcell 52.73 & \gcell 29.62 & \gcell 53.86 & \gcell 33.66 & \gcell 44.38 \\
    & \bcell VFA & \bcell 42.95 & \bcell 45.98 & \bcell 60.75 & \bcell 53.22 & \bcell 41.00 & \bcell 57.91 & \bcell 19.72 & \bcell 45.93  \\
    & \textit{\small $\Delta$ Gains} & \gbf{+11.35} & \gbf{+0.63} & \rbf{--3.08} & \gbf{+0.49} & \gbf{+11.38} & \gbf{+4.05} & \rbf{--13.94} & \gbf{+1.55} \\ 
    \midrule

    \multirow{3}{*}{\shortstack[l]{LLaVA-Next-8B \\ \footnotesize \color{gray} \firstarrow Llama3-8B-inst}}
    & \gcell Base & \gcell 25.47 & \gcell 41.95 & \gcell 53.75 & \gcell 45.89 & \gcell 36.00 & \gcell 52.50 & \gcell 30.57 & \gcell 40.88 \\
    & \bcell VFA & \bcell 45.55 & \bcell 42.55 & \bcell 60.50  & \bcell 49.73 & \bcell 36.62 & \bcell 61.77 & \bcell 33.33 & \bcell 47.15 \\
    & \textit{\small $\Delta$ Gains} & \gbf{+20.08} & \gbf{+0.60} & \gbf{+6.75} & \gbf{+3.84} & \gbf{+0.62} & \gbf{+9.27} & \gbf{+2.76} & \gbf{+6.27} \\ 
    \midrule

    \multirow{3}{*}{\shortstack[l]{LLaVA-OV-1.5-8B \\ \footnotesize \color{gray} \firstarrow Qwen3-8B-base}}
    & \gcell Base & \gcell 28.85 & \gcell 50.38 & \gcell 67.84 & \gcell 68.04 & \gcell 48.25 & \gcell 70.21 & \gcell 38.37 & \gcell 53.13 \\
    & \bcell VFA & \bcell 30.43  & \bcell 51.52  & \bcell 67.25 & \bcell 67.94 & \bcell 50.00 & \bcell 67.81 & \bcell 38.00 & \bcell 53.28 \\
    & \textit{\small $\Delta$ Gains} & \gbf{+1.58} & \gbf{+1.14} & \rbf{--0.59} & \rbf{--0.10} & \gbf{+1.75} & \rbf{--2.40} & \rbf{--0.37} & \gbf{+0.15} \\ 
    \midrule

    \multirow{3}{*}{\shortstack[l]{LLaVA-OV-1.5-4B \\ \footnotesize \color{gray} \firstarrow Qwen3-4B-base}}
    & \gcell Base & \gcell 30.68 & \gcell 49.75 & \gcell 65.42 & \gcell 63.10 & \gcell 45.88 & \gcell 73.23 & \gcell 36.88 & \gcell 52.13 \\
    & \bcell VFA & \bcell 31.38  & \bcell 49.85 & \bcell 66.33 & \bcell 64.55  & \bcell 46.88  & \bcell 70.83 & \bcell 36.81 & \bcell 52.38 \\
    & \textit{\small $\Delta$ Gains} & \gbf{+0.70} & \gbf{+0.10} & \gbf{+0.91} & \gbf{+1.45} & \gbf{+1.00} & \rbf{--2.40} & \rbf{--0.07} & \gbf{+0.25} \\ 

    \bottomrule
    \end{tabular}
    }
\caption{\textbf{Multilingual Text-task Results of VFA across Various MLLMs.} \gbf{Green} and \rbf{Red} values in $\Delta$ Gains rows denote relative changes compared to the base MLLM. \firstarrow indicates the backbone of the MLLM.}
    \label{tab:text_only_ft_comparison_MLLM}

\end{table*}

%% file: tables/tab_05_pangea_comparison.tex
\begin{table*}[htbp]

    \centering    

    \setlength{\tabcolsep}{4pt} 
    \setlength{\aboverulesep}{0pt}
    \setlength{\belowrulesep}{0pt}
    \renewcommand{\arraystretch}{1.2}
    
    \resizebox{1.0\textwidth}{!}{%
    \begin{tabular}{l|l| c|cccccc|c}
        \toprule
        \textbf{Model} & \textbf{Method} & \textbf{Training Dataset} & \textbf{MaXM} & \textbf{xGQA} & \textbf{xMMMU} & \textbf{XM100} & \textbf{MaRVL} & \textbf{M3Exam} & \textbf{Avg.} \\
        \midrule
        Pangea-7B & \multicolumn{1}{c|}{--} & 6M Multimodal Samples & 51.27 & 62.58 & 44.00 & 16.72 & 78.33 & 49.15 & 50.34 \\
        \midrule
        \multirow{3}{*}{\shortstack[l]{Qwen2-VL-7B}}
         & \gcell Base & \gcell -- & \gcell 45.03 & \gcell 42.64 & \gcell 47.97 & \gcell 2.05 & \gcell 57.83 & \gcell 57.43 & \gcell 42.16 \\
         & \bcell VFA  & \bcell 100K Text-Only Samples & \bcell 48.68 & \bcell 51.64 & \bcell 48.45 & \bcell 13.86 & \bcell 66.67 & \bcell 57.76 & \bcell 47.84 \\
         & \textit{\small $\Delta$ Gains} & -- & \gbf{+3.65} & \gbf{+9.00} & \gbf{+0.48} & \gbf{+11.81} & \gbf{+8.84} & \gbf{+0.33} & \gbf{+5.68} \\
        \bottomrule
    \end{tabular}%
    }
\caption{\textbf{Comparison with Multimodal Fine-tuning on Multilingual Multimodal Benchmarks.} We compare VFA (trained on limited text-only data) against a baseline MLLM trained with large-scale image-text supervision, highlighting the trade-off between performance and training efficiency.}    \label{tab:main_multimodal_results}
    \vspace{-10pt}
\end{table*}

%% file: tables/tab_06_visual_coding_results.tex
\begin{table}[t]
    \centering
    
    \setlength{\tabcolsep}{1.2mm}
\resizebox{1.0\linewidth}{!}{
    \large
    \begin{tabular}{l|l|cc|c}
    \toprule
    \textbf{Model} & \textbf{Method} & \textbf{CodeVision} & \textbf{HumanEval-V} & \textbf{Avg.} \\ \hline
    \multirow{3}{*}{InternVL3-8B} & \gcell Base & \gcell 39.96 & \gcell 18.28 & \gcell 29.12 \\
     & \bcell VFA & \bcell 47.49 & \bcell 19.93 & \bcell 33.71 \\
     & \textit{\small $\Delta$ Gains} & \gbf{+7.53} & \gbf{+1.65} & \gbf{+4.59} \\ \hline
    \multirow{3}{*}{Llama3-LLaVA-8B} & \gcell Base & \gcell 8.35 & \gcell 6.04 & \gcell 7.20 \\
     & \bcell VFA & \bcell 8.43 & \bcell 6.61 & \bcell 7.52 \\
     & \textit{\small $\Delta$ Gains} & \gbf{+0.08} & \gbf{+0.57} & \gbf{+0.32} \\ \hline
    \multirow{3}{*}{Idefics3-8B} & \gcell Base & \gcell 3.67 & \gcell 9.69 & \gcell 6.68 \\
     & \bcell VFA & \bcell 15.58 & \bcell 11.01 & \bcell 13.30 \\
     & \textit{\small $\Delta$ Gains} & \gbf{+11.91} & \gbf{+1.32} & \gbf{+6.62} \\
    \bottomrule
    \end{tabular}
    }
    \caption{\textbf{VFA on Multimodal Visual Coding Tasks}.}
        \label{tab:VFA_summary_large_font}

    \vspace{-5pt}

\end{table}

%% file: tables/tab_07_language_codes.tex
\begin{table*}[htbp]
\centering
\small
\setlength{\tabcolsep}{6pt} 
\renewcommand{\arraystretch}{1.1}
\begin{tabular}{ll @{\hspace{2.5em}} ll @{\hspace{2.5em}} ll}
    \toprule
    \textbf{Code} & \textbf{Language} & \textbf{Code} & \textbf{Language} & \textbf{Code} & \textbf{Language} \\
    \midrule
    af & Afrikaans & it & Italian    & su & Sundanese \\
    ar & Arabic    & iw (he) & Hebrew  & sw & Swahili \\
    bn & Bengali   & ja & Japanese   & ta & Tamil \\
    de & German    & jv & Javanese   & th & Thai \\
    en & English   & ko & Korean     & tr & Turkish \\
    es & Spanish   & mn & Mongolian  & ur & Urdu \\
    fr & French    & pt & Portuguese & vi & Vietnamese \\
    hi & Hindi     & ro & Romanian   & zh & Chinese \\
    id & Indonesian& ru & Russian    & min & Minangkabau \\
    rw & Kinyarwanda & -- & -- & -- & -- \\
    \bottomrule
\end{tabular}
\caption{Mapping between ISO 639-1 language codes and full names.}
\label{tab:language_codes}
\end{table*}


%% file: tables/tab_08_training_efficiency.tex
\begin{table*}[!htbp]
    \centering
    \small
    \renewcommand{\arraystretch}{1.2} 
    \begin{tabular}{@{} l c c c @{}}
        \toprule
        \textbf{Method} & \textbf{Training Data} & \textbf{GPU Hours (8B)} & \textbf{Throughput} \\
        \midrule
        Multimodal SFT & 760K image-text & $\sim$320 A100-h & $\sim$2.4K/GPU-h \\
        VFA    & 100K text-only  & $\sim$10 A100-h  & $\sim$10K/GPU-h \\
        \bottomrule
    \end{tabular}
    \caption{Comparison of training efficiency between standard multimodal SFT and VFA.}
    \label{tab:efficiency_comparison}
\end{table*}

%% file: tables/tab_09_p_mmeval_benchmark.tex
\begin{table*}[!htbp]
    \centering
\small 

    \setlength{\tabcolsep}{4pt} 
    \setlength{\aboverulesep}{0pt}
    \setlength{\belowrulesep}{0pt}
    \renewcommand{\arraystretch}{1.2}

\setlength{\tabcolsep}{10pt} 
\renewcommand{\arraystretch}{1.2}

\begin{tabularx}{\linewidth}{l X l}
\toprule
\textbf{Task} & \textbf{Benchmarks} & \textbf{Metric} \\
\midrule
Machine Translation & FLORES-200~\citep{nllb2022} & BLEU \\
Natural Language Understanding & XNLI~\citep{conneau2018xnli}, M-HellaSwag~\citep{lai2023okapi} & Accuracy \\
Code Generation & HumanEval-XL~\citep{peng2024humaneval} & Pass@1 \\
Mathematical Reasoning & MGSM~\citep{shilanguage} & Accuracy \\
Logical Reasoning & M-LogiQA~\citep{liu2021logiqa} & Accuracy \\
General Knowledge & M-MMLU~\citep{hendryckstest2021} & Accuracy \\
Instruction Following & M-IFEval~\citep{zhou2023instruction} & Accuracy \\
\bottomrule
\end{tabularx}
\caption{\textbf{Overview of the P-MMEval Benchmark.} Summary of evaluation tasks, multilingual benchmarks, and their corresponding metrics. This suite evaluates the core linguistic and reasoning capabilities of models across diverse languages.}
\label{tab:P-MMEval-benchmarks}
\end{table*}

%% file: tables/tab_10_multimodal_benchmarks.tex
\begin{table*}[htbp]
    \centering
    \small

    \setlength{\tabcolsep}{4pt} 
    \setlength{\aboverulesep}{0pt}
    \setlength{\belowrulesep}{0pt}
    \renewcommand{\arraystretch}{1.2}
    
    \setlength{\tabcolsep}{4pt}
    \renewcommand{\arraystretch}{1.2} 
    
    \begin{tabularx}{\linewidth}{l >{\raggedright\arraybackslash}p{4.7cm} c c X c}
        \toprule
        \textbf{Tasks} & \textbf{Datasets} & \textbf{Forms} & \textbf{Size} & \textbf{Languages} & \textbf{Metric} \\
        \midrule
        
        \multirow{2.5}{*}{\shortstack[l]{Multimodal\\Chat}} 
          & xChatBench~\citep{yue2024pangea} & Long & 400 & zh, en, hi, id, ja, rw, ko, es & LLM-as-Judge \\
          & M-LlavaBench~\citep{PALO} & Long & 600 & ar, bn, zh, fr, hi, ja, ru, es, ur, en & LLM-as-Judge \\
        \midrule
        
        Captioning & XM100~\citep{yue2024pangea} & Long & 3.6K & 36 languages & ROUGE-L \\
        \midrule
        
        \multirow{2.5}{*}{\shortstack[l]{Cultural\\Understanding}} 
          & CVQA~\citep{mogrovejo2024cvqa} & MC & 21K & en, zh, ko, mn, ja, id, jv, min, su & Accuracy \\
          & MaRVL~\citep{liu-etal-2021-visually} & Short & 6K & id, sw, ta, tr, zh & Accuracy \\
        \midrule
        
        \multirow{2.5}{*}{\shortstack[l]{Multilingual\\VQA}} 
          & xGQA~\citep{pfeiffer-etal-2022-xgqa} & Short & 77K & en, de, pt, ru, id, bn, ko, zh & Accuracy \\
          & MaXM~\citep{changpinyo-etal-2023-maxm} & MC & 2K & hi, th, zh, fr, en, iw, ro & Accuracy \\
        \midrule
        
        \multirow{2.5}{*}{\shortstack[l]{Reasoning\\(Multi-subject)}} 
          & xMMMU~\citep{yue2024pangea} & Short/MC & 3K & en, ar, fr, hi, id, ja, pt & Accuracy \\
          & M3Exam~\citep{zhang2023mexam} & MC & 3K & en, zh, it, pt, vi, th, af & Accuracy \\
        
        \bottomrule
    \end{tabularx}
    \caption{\textbf{Summary of Evaluation Datasets.} We select CVQA for model validation. MaRVL is utilized for multilingual visual reasoning, serving as an extension of the NLVR2 task. The table lists task categories, dataset forms, language coverage, and metrics.}
    \label{tab:multimodal_benchmark}
\end{table*}

%% file: tables/tab_11_model_backbones.tex
\begin{table}
    \centering
    \small
    \renewcommand{\arraystretch}{1.2}
    \resizebox{\columnwidth}{!}{
    \begin{tabular}{@{} lll @{}} 
        \toprule
        \textbf{Family} & \textbf{MLLM} & \textbf{Base LLM} \\
        \midrule
        \multirow{4}{*}{Qwen} 
        & Qwen2-VL-7B-Instruct & Qwen2-7B \\
        & Qwen2.5-VL-7B-Instruct & Qwen2.5-7B \\
        & LLaVA-OV-1.5-4B-Inst & Qwen3-4B-Base \\
        & LLaVA-OV-1.5-8B-Inst & Qwen3-8B-Base \\
        \midrule
        \multirow{2}{*}{Llama} 
        & llama3-llava-next-8b-hf & Meta-Llama-3-8B-Inst \\
        & Idefics3-8B-Llama3 & Llama-3.1-8B-Inst \\
        \bottomrule
    \end{tabular}
    }
    \caption{Multimodal models and their corresponding LLM backbones used in this paper.}
    \label{tab:model_specifications}
\end{table}

%% file: tables/tab_12_selected_merge_configs.tex
\begin{table}[!t]
    \centering 
    \small
    \renewcommand{\arraystretch}{1.2}
    \setlength{\tabcolsep}{4pt}
    \setlength{\aboverulesep}{0pt}
    \setlength{\belowrulesep}{0pt}

    \begin{tabular}{lcc}
        \toprule
        \textbf{Model} & \textbf{Merge Operator} & \boldmath$\alpha$ \\
        \midrule
        Qwen2.5-VL-7B   & WA   & 0.9 \\
        Idefics3-8B     & TIES & 0.9 \\
        LLaVA-Next-8B   & TA   & 0.5 \\
        LLaVA-OV-1.5-8B & TIES   & 1.0 \\
        LLaVA-OV-1.5-4B & WA   & 0.9 \\
        \bottomrule
    \end{tabular}

    \caption{Selected merge operator and mixing coefficient ($\alpha$) per model based on CVQA validation.}
    \label{tab:selected_merge_configs}
\end{table}

%% file: tables/tab_13_maxm_results.tex
\begin{table*}[t]
    \centering
    \small

    \setlength{\aboverulesep}{0pt}
    \setlength{\belowrulesep}{0pt}
    \renewcommand{\arraystretch}{1.2}
    \setlength{\tabcolsep}{8pt}

\resizebox{1.0\textwidth}{!}{
\begin{tabular}{l|l|ccccccc} 
\toprule
\textbf{Model} & \textbf{Method} & \textbf{English} & \textbf{Thai} & \textbf{Chinese} & \textbf{French} & \textbf{Hindi} & \textbf{Hebrew} & \textbf{Romanian} \\
\midrule

\multirow{3}{*}{Qwen2.5-VL-7B}
 & \gcell Base & \gcell 57.26 & \gcell 64.55 & \gcell 46.57 & \gcell 46.97 & \gcell 51.15 & \gcell 49.64 & \gcell 37.68 \\
 & \bcell VFA & \bcell 56.42 & \bcell 65.30 & \bcell 49.82 & \bcell 50.00 & \bcell 51.15 & \bcell 49.29 & \bcell 40.85 \\
 & \textit{\small $\Delta$ Gains} & \rbf{--0.84} & \gbf{+0.75} & \gbf{+3.25} & \gbf{+3.03} & \gbf{+0.00} & \rbf{--0.35} & \gbf{+3.17} \\
 \midrule

\multirow{3}{*}{Idefics3-8B}
 & \gcell Base & \gcell 55.25 & \gcell 44.78 & \gcell 36.46 & \gcell 43.56 & \gcell 61.54 & \gcell 46.43 & \gcell 38.38 \\
 & \bcell VFA & \bcell 52.14 & \bcell 56.72 & \bcell 39.71 & \bcell 44.32 & \bcell 63.08 & \bcell 54.29 & \bcell 41.55 \\
 & \textit{\small $\Delta$ Gains} & \rbf{--3.11} & \gbf{+11.94} & \gbf{+3.25} & \gbf{+0.76} & \gbf{+1.54} & \gbf{+7.86} & \gbf{+3.17} \\
 \midrule

\multirow{3}{*}{LLaVA-Next-8B}
 & \gcell Base & \gcell 49.81 & \gcell 33.58 & \gcell 29.60 & \gcell 33.33 & \gcell 19.62 & \gcell 20.71 & \gcell 26.41 \\
 & \bcell VFA & \bcell 48.64 & \bcell 53.36 & \bcell 41.88 & \bcell 40.53 & \bcell 41.54 & \bcell 37.14 & \bcell 37.32 \\
 & \textit{\small $\Delta$ Gains} & \rbf{--1.17} & \gbf{+19.78} & \gbf{+12.28} & \gbf{+7.20} & \gbf{+21.92} & \gbf{+16.43} & \gbf{+10.91} \\
 \midrule

\multirow{3}{*}{LLaVA-OV-1.5-8B}
 & \gcell Base & \gcell 55.25 & \gcell 66.04 & \gcell 49.10 & \gcell 61.74 & \gcell 56.54 & \gcell 48.57 & \gcell 41.90 \\
 & \bcell VFA & \bcell 60.70 & \bcell 68.28 & \bcell 50.90 & \bcell 62.88 & \bcell 58.85 & \bcell 46.79 & \bcell 46.13 \\
 & \textit{\small $\Delta$ Gains} & \gbf{+5.45} & \gbf{+2.24} & \gbf{+1.80} & \gbf{+1.14} & \gbf{+2.31} & \rbf{--1.78} & \gbf{+4.23} \\
 \midrule

\multirow{3}{*}{LLaVA-OV-1.5-4B}
 & \gcell Base & \gcell 57.59 & \gcell 61.94 & \gcell 41.52 & \gcell 56.82 & \gcell 49.23 & \gcell 40.36 & \gcell 40.85 \\
 & \bcell VFA & \bcell 62.65 & \bcell 63.06 & \bcell 48.38 & \bcell 56.44 & \bcell 51.92 & \bcell 41.07 & \bcell 39.79 \\
 & \textit{\small $\Delta$ Gains} & \gbf{+5.06} & \gbf{+1.12} & \gbf{+6.86} & \rbf{--0.38} & \gbf{+2.69} & \gbf{+0.71} & \rbf{--1.06} \\
\bottomrule
\end{tabular}
}
\caption{Performance on MAXM across multiple languages.}
\label{tab:maxm_results}

\end{table*}

%% file: tables/tab_14_xgqa_results.tex
\begin{table*}[t]

    \setlength{\tabcolsep}{4pt} 
    \setlength{\aboverulesep}{0pt}
    \setlength{\belowrulesep}{0pt}
    \renewcommand{\arraystretch}{1.2}

\centering
\setlength{\tabcolsep}{6pt} 
\resizebox{1.0\textwidth}{!}{
\begin{tabular}{l|l| cccccccc}
\toprule
\textbf{Model} & \textbf{Method} & \textbf{Bengali} & \textbf{German} & \textbf{English} & \textbf{Indonesian} & \textbf{Korean} & \textbf{Portuguese} & \textbf{Russian} & \textbf{Chinese} \\
\midrule

\multirow{3}{*}{Qwen2.5-VL-7B}
 & \gcell Base & \gcell 44.60 & \gcell 47.70 & \gcell 64.70 & \gcell 43.20 & \gcell 45.60 & \gcell 48.30 & \gcell 49.70 & \gcell 38.70 \\
 & \bcell VFA      & \bcell 45.10 & \bcell 49.50 & \bcell 63.00 & \bcell 44.30 & \bcell 46.40 & \bcell 48.60 & \bcell 50.80 & \bcell 39.90 \\
 & \textit{\small $\Delta$ Gains} & \gbf{+0.50} & \gbf{+1.80} & \rbf{--1.70} & \gbf{+1.10} & \gbf{+0.80} & \gbf{+0.30} & \gbf{+1.10} & \gbf{+1.20} \\ 
 \midrule

\multirow{3}{*}{Idefics3-8B}
 & \gcell Base & \gcell 39.80 & \gcell 50.20 & \gcell 57.90 & \gcell 46.00 & \gcell 48.10 & \gcell 48.20 & \gcell 48.80 & \gcell 50.80 \\
 & \bcell VFA      & \bcell 44.20 & \bcell 50.00 & \bcell 56.30 & \bcell 46.20 & \bcell 46.90 & \bcell 47.10 & \bcell 46.60 & \bcell 47.50 \\
 & \textit{\small $\Delta$ Gains} & \gbf{+4.40} & \rbf{--0.20} & \rbf{--1.60} & \gbf{+0.20} & \rbf{--1.20} & \rbf{--1.10} & \rbf{--2.20} & \rbf{--3.30} \\ 
 \midrule

\multirow{3}{*}{LLaVA-Next-8B}
 & \gcell Base & \gcell 12.20 & \gcell 43.30 & \gcell 68.20 & \gcell 40.00 & \gcell 43.20 & \gcell 47.00 & \gcell 45.20 & \gcell 49.70 \\
 & \bcell VFA      & \bcell 17.20 & \bcell 49.20 & \bcell 63.80 & \bcell 41.40 & \bcell 44.70 & \bcell 50.20 & \bcell 48.00 & \bcell 48.60 \\
 & \textit{\small $\Delta$ Gains} & \gbf{+5.00} & \gbf{+5.90} & \rbf{--4.40} & \gbf{+1.40} & \gbf{+1.50} & \gbf{+3.20} & \gbf{+2.80} & \rbf{--1.10} \\ 
 \midrule

\multirow{3}{*}{\shortstack[l]{LLaVA-OV-1.5-8B}}
 & \gcell Base & \gcell 40.60 & \gcell 30.70 & \gcell 64.50 & \gcell 28.10 & \gcell 17.60 & \gcell 27.80 & \gcell 29.50 & \gcell 0.00 \\
 & \bcell VFA      & \bcell 45.50 & \bcell 52.40 & \bcell 62.20 & \bcell 49.40 & \bcell 42.20 & \bcell 52.20 & \bcell 47.10 & \bcell 0.80 \\
 & \textit{\small $\Delta$ Gains} & \gbf{+4.90} & \gbf{+21.70} & \rbf{--2.30} & \gbf{+21.30} & \gbf{+24.60} & \gbf{+24.40} & \gbf{+17.60} & \gbf{+0.80} \\ 
 \midrule

\multirow{3}{*}{\shortstack[l]{LLaVA-OV-1.5-4B}}
 & \gcell Base & \gcell 37.40 & \gcell 51.00 & \gcell 64.70 & \gcell 40.00 & \gcell 18.60 & \gcell 44.50 & \gcell 35.60 & \gcell 0.10 \\
 & \bcell VFA      & \bcell 38.40 & \bcell 51.30 & \bcell 64.30 & \bcell 40.70 & \bcell 23.70 & \bcell 44.60 & \bcell 38.70 & \bcell 0.00 \\
 & \textit{\small $\Delta$ Gains} & \gbf{+1.00} & \gbf{+0.30} & \rbf{--0.40} & \gbf{+0.70} & \gbf{+5.10} & \gbf{+0.10} & \gbf{+3.10} & \rbf{--0.10} \\
\bottomrule
\end{tabular}
}
\caption{Performance on XGQA across multiple languages.}
\label{tab:xgqa_results}

\end{table*}

%% file: tables/tab_15_xmmmu_results.tex
\begin{table*}[ht]
    \centering
    \small

    \setlength{\aboverulesep}{0pt}
    \setlength{\belowrulesep}{0pt}
    \renewcommand{\arraystretch}{1.2}
    \setlength{\tabcolsep}{6pt} 

\resizebox{1.0\textwidth}{!}{
\begin{tabular}{l|l|ccccccc} 
\toprule
\textbf{Model} & \textbf{Method} & \textbf{Arabic} & \textbf{English} & \textbf{French} & \textbf{Hindi} & \textbf{Indonesian} & \textbf{Japanese} & \textbf{Portuguese} \\
\midrule

\multirow{3}{*}{Qwen2.5-VL-7B}
 & \gcell Base & \gcell 42.60 & \gcell 50.20 & \gcell 50.70 & \gcell 44.00 & \gcell 49.20 & \gcell 46.10 & \gcell 51.50 \\
 & \bcell VFA & \bcell 43.30 & \bcell 51.10 & \bcell 48.70 & \bcell 42.60 & \bcell 48.10 & \bcell 46.80 & \bcell 50.20 \\
 & \textit{\small $\Delta$ Gains} & \gbf{+0.70} & \gbf{+0.90} & \rbf{--2.00} & \rbf{--1.40} & \rbf{--1.10} & \gbf{+0.70} & \rbf{--1.30} \\
 \midrule

\multirow{3}{*}{Idefics3-8B}
 & \gcell Base & \gcell 37.20 & \gcell 42.70 & \gcell 44.60 & \gcell 42.60 & \gcell 43.40 & \gcell 46.10 & \gcell 43.80 \\
 & \bcell VFA & \bcell 38.60 & \bcell 44.40 & \bcell 44.00 & \bcell 43.60 & \bcell 39.70 & \bcell 43.50 & \bcell 46.80 \\
 & \textit{\small $\Delta$ Gains} & \gbf{+1.40} & \gbf{+1.70} & \rbf{--0.60} & \gbf{+1.00} & \rbf{--3.70} & \rbf{--2.60} & \gbf{+3.00} \\
 \midrule

\multirow{3}{*}{LLaVA-Next-8B}
 & \gcell Base & \gcell 38.90 & \gcell 38.40 & \gcell 41.30 & \gcell 34.70 & \gcell 35.40 & \gcell 36.10 & \gcell 37.70 \\
 & \bcell VFA & \bcell 35.90 & \bcell 41.70 & \bcell 39.60 & \bcell 34.00 & \bcell 36.40 & \bcell 35.70 & \bcell 40.40 \\
 & \textit{\small $\Delta$ Gains} & \rbf{--3.00} & \gbf{+3.30} & \rbf{--1.70} & \rbf{--0.70} & \gbf{+1.00} & \rbf{--0.40} & \gbf{+2.70} \\
 \midrule

\multirow{3}{*}{LLaVA-OV-1.5-8B}
 & \gcell Base & \gcell 51.30 & \gcell 56.10 & \gcell 55.40 & \gcell 49.80 & \gcell 58.60 & \gcell 52.40 & \gcell 56.20 \\
 & \bcell VFA & \bcell 53.00 & \bcell 55.60 & \bcell 53.70 & \bcell 47.40 & \bcell 57.60 & \bcell 50.90 & \bcell 53.20 \\
 & \textit{\small $\Delta$ Gains} & \gbf{+1.70} & \rbf{--0.50} & \rbf{--1.70} & \rbf{--2.40} & \rbf{--1.00} & \rbf{--1.50} & \rbf{--3.00} \\
 \midrule

\multirow{3}{*}{LLaVA-OV-1.5-4B}
 & \gcell Base & \gcell 52.30 & \gcell 53.80 & \gcell 56.70 & \gcell 46.40 & \gcell 53.50 & \gcell 53.90 & \gcell 56.60 \\
 & \bcell VFA & \bcell 51.30 & \bcell 54.20 & \bcell 55.00 & \bcell 47.10 & \bcell 54.20 & \bcell 54.30 & \bcell 56.20 \\
 & \textit{\small $\Delta$ Gains} & \rbf{--1.00} & \gbf{+0.40} & \rbf{--1.70} & \gbf{+0.70} & \gbf{+0.70} & \gbf{+0.40} & \rbf{--0.40} \\
\bottomrule
\end{tabular}
}
\caption{Performance on xMMMU (validation set split) across multiple languages.}
\label{tab:xmmmu_multilingual_val}

\end{table*}

%% file: tables/tab_16_marvl_results.tex
\begin{table*}[ht]
\centering
\small

    \setlength{\tabcolsep}{4pt} 
    \setlength{\aboverulesep}{0pt}
    \setlength{\belowrulesep}{0pt}
    \renewcommand{\arraystretch}{1.2}

\setlength{\tabcolsep}{5pt}
\begin{tabular}{l|l| cccccc}
\toprule
\textbf{Model} & \textbf{Method} & \textbf{English} & \textbf{Indonesian} & \textbf{Swahili} & \textbf{Tamil} & \textbf{Turkish} & \textbf{Chinese} \\
\midrule

\multirow{3}{*}{\shortstack[l]{Qwen2.5-VL-7B}}
 & \gcell Base & \gcell 62.00 & \gcell 67.00 & \gcell 44.00 & \gcell 67.00 & \gcell 39.00 & \gcell 42.00 \\
 & \bcell VFA & \bcell 75.00 & \bcell 71.00 & \bcell 54.00 & \bcell 67.00 & \bcell 61.00 & \bcell 67.00 \\
 & \textit{\small $\Delta$ Gains} & \gbf{+13.00} & \gbf{+4.00} & \gbf{+10.00} & \gbf{+0.00} & \gbf{+22.00} & \gbf{+25.00} \\ 
 \midrule

\multirow{3}{*}{\shortstack[l]{Idefics3-8B}}
 & \gcell Base & \gcell 49.00 & \gcell 18.00 & \gcell 25.00 & \gcell 19.00 & \gcell 23.00 & \gcell 25.00 \\
 & \bcell VFA & \bcell 71.00 & \bcell 59.00 & \bcell 58.00 & \bcell 61.00 & \bcell 70.00 & \bcell 57.00 \\
 & \textit{\small $\Delta$ Gains} & \gbf{+22.00} & \gbf{+41.00} & \gbf{+33.00} & \gbf{+42.00} & \gbf{+47.00} & \gbf{+32.00} \\ 
 \midrule

\multirow{3}{*}{\shortstack[l]{LLaVA-Next-8B}}
 & \gcell Base & \gcell 31.00 & \gcell 54.00 & \gcell 52.00 & \gcell 45.00 & \gcell 42.00 & \gcell 54.00 \\
 & \bcell VFA & \bcell 54.00 & \bcell 49.00 & \bcell 56.00 & \bcell 54.00 & \bcell 56.00 & \bcell 63.00 \\
 & \textit{\small $\Delta$ Gains} & \gbf{+23.00} & \rbf{--5.00} & \gbf{+4.00} & \gbf{+9.00} & \gbf{+14.00} & \gbf{+9.00} \\ 
 \midrule

\multirow{3}{*}{\shortstack[l]{LLaVA-OV-1.5-8B}}
 & \gcell Base & \gcell 67.00 & \gcell 54.00 & \gcell 51.00 & \gcell 60.00 & \gcell 74.00 & \gcell 67.00 \\
 & \bcell VFA & \bcell 69.00 & \bcell 65.00 & \bcell 52.00 & \bcell 63.00 & \bcell 73.00 & \bcell 71.00 \\
 & \textit{\small $\Delta$ Gains} & \gbf{+2.00} & \gbf{+11.00} & \gbf{+1.00} & \gbf{+3.00} & \rbf{--1.00} & \gbf{+4.00} \\ 
 \midrule

\multirow{3}{*}{\shortstack[l]{LLaVA-OV-1.5-4B}}
 & \gcell Base & \gcell 65.00 & \gcell 59.00 & \gcell 50.00 & \gcell 61.00 & \gcell 66.00 & \gcell 62.00 \\
 & \bcell VFA & \bcell 64.00 & \bcell 60.00 & \bcell 50.00 & \bcell 61.00 & \bcell 67.00 & \bcell 62.00 \\
 & \textit{\small $\Delta$ Gains} & \rbf{--1.00} & \gbf{+1.00} & \gbf{+0.00} & \gbf{+0.00} & \gbf{+1.00} & \gbf{+0.00} \\
\bottomrule
\end{tabular}
\caption{Performance on MaRVL across multiple languages.}
\label{tab:marvl_results}

\end{table*}

%% file: tables/tab_17_m3exam_results.tex
\begin{table*}[ht]
    \centering
    \small

    \setlength{\aboverulesep}{0pt}
    \setlength{\belowrulesep}{0pt}
    \renewcommand{\arraystretch}{1.2}
    \setlength{\tabcolsep}{4pt}

\resizebox{1.0\textwidth}{!}{
\begin{tabular}{l|l|ccccccc} 
    \toprule
    \textbf{Model} & \textbf{Method} & \textbf{Afrikaans} & \textbf{Chinese} & \textbf{English} & \textbf{Italian} & \textbf{Portuguese} & \textbf{Thai} & \textbf{Vietnamese} \\
    \midrule

    \multirow{3}{*}{Qwen2.5-VL-7B}
     & \gcell Base & \gcell 60.74 & \gcell 86.37 & \gcell 67.09 & \gcell 67.00 & \gcell 50.44 & \gcell 40.40 & \gcell 39.66 \\
     & \bcell VFA & \bcell 61.96 & \bcell 86.14 & \bcell 67.47 & \bcell 67.00 & \bcell 52.67 & \bcell 40.40 & \bcell 41.38 \\
     & \textit{\small $\Delta$ Gains} & \gbf{+1.22} & \rbf{--0.23} & \gbf{+0.38} & \gbf{+0.00} & \gbf{+2.23} & \gbf{+0.00} & \gbf{+1.72} \\
     \midrule

    \multirow{3}{*}{Idefics3-8B}
     & \gcell Base & \gcell 34.97 & \gcell 22.40 & \gcell 23.33 & \gcell 48.87 & \gcell 24.00 & \gcell 23.94 & \gcell 27.59 \\
     & \bcell VFA & \bcell 56.44 & \bcell 54.27 & \bcell 55.61 & \bcell 54.66 & \bcell 46.00 & \bcell 34.66 & \bcell 32.76 \\
     & \textit{\small $\Delta$ Gains} & \gbf{+21.47} & \gbf{+31.87} & \gbf{+32.28} & \gbf{+5.79} & \gbf{+22.00} & \gbf{+10.72} & \gbf{+5.17} \\
     \midrule

    \multirow{3}{*}{LLaVA-Next-8B}
     & \gcell Base & \gcell 42.94 & \gcell 48.04 & \gcell 53.72 & \gcell 51.13 & \gcell 39.33 & \gcell 31.92 & \gcell 34.48 \\
     & \bcell VFA & \bcell 41.10 & \bcell 48.27 & \bcell 55.61 & \bcell 52.64 & \bcell 42.00 & \bcell 31.17 & \bcell 36.21 \\
     & \textit{\small $\Delta$ Gains} & \rbf{--1.84} & \gbf{+0.23} & \gbf{+1.89} & \gbf{+1.51} & \gbf{+2.67} & \rbf{--0.75} & \gbf{+1.73} \\
     \midrule

    \multirow{3}{*}{LLaVA-OV-1.5-8B-Inst}
     & \gcell Base & \gcell 69.94 & \gcell 73.67 & \gcell 69.86 & \gcell 70.78 & \gcell 54.22 & \gcell 45.39 & \gcell 55.17 \\
     & \bcell VFA & \bcell 69.33 & \bcell 74.83 & \bcell 69.99 & \bcell 70.53 & \bcell 54.67 & \bcell 44.14 & \bcell 51.72 \\
     & \textit{\small $\Delta$ Gains} & \rbf{--0.61} & \gbf{+1.16} & \gbf{+0.13} & \rbf{--0.25} & \gbf{+0.45} & \rbf{--1.25} & \rbf{--3.45} \\
     \midrule

    \multirow{3}{*}{LLaVA-OV-1.5-4B-Inst}
     & \gcell Base & \gcell 63.80 & \gcell 66.28 & \gcell 66.71 & \gcell 67.00 & \gcell 51.56 & \gcell 43.14 & \gcell 47.41 \\
     & \bcell VFA & \bcell 66.26 & \bcell 66.51 & \bcell 67.34 & \bcell 65.24 & \bcell 52.44 & \bcell 45.14 & \bcell 45.69 \\
     & \textit{\small $\Delta$ Gains} & \gbf{+2.46} & \gbf{+0.23} & \gbf{+0.63} & \rbf{--1.76} & \gbf{+0.88} & \gbf{+2.00} & \rbf{--1.72} \\
    \bottomrule
\end{tabular}
}
    \caption{Performance of different MLLMs on M3Exam across multiple languages.}
    \label{tab:m3exam_results}
\end{table*}

%% file: tables/tab_18_xm100_results.tex
\begin{table*}[h!]
\centering

    \setlength{\tabcolsep}{4pt}
    \setlength{\aboverulesep}{0pt}
    \setlength{\belowrulesep}{0pt}
    \renewcommand{\arraystretch}{1.25}

\resizebox{1.0\textwidth}{!}{%
\begin{tabular}{l|l| ccccccccc}
\toprule
\textbf{Model} & \textbf{Method} & \textbf{Arabic} & \textbf{Bengali} & \textbf{Czech} & \textbf{Danish} & \textbf{German} & \textbf{Greek} & \textbf{English} & \textbf{Spanish} & \textbf{Persian} \\
\midrule
\multirow{3}{*}{Qwen2.5-VL-7B}
 & \gcell Base & \gcell 11.04 & \gcell 12.74 & \gcell 16.65 & \gcell 22.24 & \gcell 17.69 & \gcell 12.07 & \gcell 25.30 & \gcell 23.94 & \gcell 20.57 \\
 & \bcell VFA      & \bcell 11.38 & \bcell 10.49 & \bcell 15.43 & \bcell 20.82 & \bcell 16.61 & \bcell 12.67 & \bcell 28.34 & \bcell 24.02 & \bcell 22.02 \\
 & \textit{\small $\Delta$ Gains} & \gbf{+0.34} & \rbf{--2.25} & \rbf{--1.22} & \rbf{--1.42} & \rbf{--1.08} & \gbf{+0.60} & \gbf{+3.04} & \gbf{+0.08} & \gbf{+1.45} \\
 \midrule

\multirow{3}{*}{Idefics3-8B}
 & \gcell Base & \gcell 11.65 & \gcell 15.59 & \gcell 13.15 & \gcell 20.34 & \gcell 14.35 & \gcell 11.13 & \gcell 23.86 & \gcell 21.10 & \gcell 16.09 \\
 & \bcell VFA      & \bcell 12.30 & \bcell 16.07 & \bcell 12.94 & \bcell 22.99 & \bcell 17.87 & \bcell 13.66 & \bcell 30.37 & \bcell 24.31 & \bcell 20.95 \\
 & \textit{\small $\Delta$ Gains} & \gbf{+0.65} & \gbf{+0.48} & \rbf{--0.21} & \gbf{+2.65} & \gbf{+3.52} & \gbf{+2.53} & \gbf{+6.51} & \gbf{+3.21} & \gbf{+4.86} \\
 \midrule

\multirow{3}{*}{LLaVA-Next-8B}
 & \gcell Base & \gcell 0.00 & \gcell 0.00 & \gcell 1.98 & \gcell 1.04 & \gcell 2.05 & \gcell 0.00 & \gcell 30.38 & \gcell 1.48 & \gcell 0.00 \\
 & \bcell VFA      & \bcell 8.28 & \bcell 8.79 & \bcell 13.58 & \bcell 23.23 & \bcell 17.65 & \bcell 7.30 & \bcell 31.04 & \bcell 26.18 & \bcell 21.79 \\
 & \textit{\small $\Delta$ Gains} & \gbf{+8.28} & \gbf{+8.79} & \gbf{+11.60} & \gbf{+22.19} & \gbf{+15.60} & \gbf{+7.30} & \gbf{+0.66} & \gbf{+24.70} & \gbf{+21.79} \\
 \midrule

\textbf{Model} & \textbf{Method} & \textbf{Finnish} & \textbf{Filipino} & \textbf{French} & \textbf{Hebrew} & \textbf{Hindi} & \textbf{Croatian} & \textbf{Hungarian} & \textbf{Indonesian} & \textbf{Italian} \\
\midrule
\multirow{3}{*}{Qwen2.5-VL-7B}
 & \gcell Base & \gcell 9.29 & \gcell 21.13 & \gcell 23.78 & \gcell 10.66 & \gcell 15.82 & \gcell 13.41 & \gcell 7.16 & \gcell 24.45 & \gcell 20.86 \\
 & \bcell VFA      & \bcell 9.04 & \bcell 20.60 & \bcell 21.82 & \bcell 12.09 & \bcell 14.99 & \bcell 12.59 & \bcell 16.09 & \bcell 22.06 & \bcell 20.36 \\
 & \textit{\small $\Delta$ Gains} & \rbf{--0.25} & \rbf{--0.53} & \rbf{--1.96} & \gbf{+1.43} & \rbf{--0.83} & \rbf{--0.82} & \gbf{+8.93} & \rbf{--2.39} & \rbf{--0.50} \\
 \midrule

\multirow{3}{*}{Idefics3-8B}
 & \gcell Base & \gcell 6.26 & \gcell 20.11 & \gcell 19.03 & \gcell 8.76 & \gcell 12.94 & \gcell 9.93 & \gcell 13.02 & \gcell 22.97 & \gcell 17.28 \\
 & \bcell VFA      & \bcell 8.84 & \bcell 23.71 & \bcell 23.45 & \bcell 13.97 & \bcell 15.72 & \bcell 11.64 & \bcell 15.38 & \bcell 25.84 & \bcell 22.11 \\
 & \textit{\small $\Delta$ Gains} & \gbf{+2.58} & \gbf{+3.60} & \gbf{+4.42} & \gbf{+5.21} & \gbf{+2.78} & \gbf{+1.71} & \gbf{+2.36} & \gbf{+2.87} & \gbf{+4.83} \\
 \midrule
\multirow{3}{*}{LLaVA-Next-8B}
 & \gcell Base & \gcell 0.00 & \gcell 0.39 & \gcell 0.91 & \gcell 1.71 & \gcell 0.00 & \gcell 0.00 & \gcell 0.07 & \gcell 3.25 & \gcell 0.67 \\
 & \bcell VFA      & \bcell 9.56 & \bcell 18.49 & \bcell 25.72 & \bcell 11.42 & \bcell 13.96 & \bcell 8.86 & \bcell 14.36 & \bcell 19.12 & \bcell 22.32 \\
 & \textit{\small $\Delta$ Gains} & \gbf{+9.56} & \gbf{+18.10} & \gbf{+24.81} & \gbf{+9.71} & \gbf{+13.96} & \gbf{+8.86} & \gbf{+14.29} & \gbf{+15.87} & \gbf{+21.65} \\
 \midrule

\textbf{Model} & \textbf{Method} & \textbf{Japanese} & \textbf{Korean} & \textbf{Maori} & \textbf{Dutch} & \textbf{Norwegian} & \textbf{Polish} & \textbf{Portuguese} & \textbf{Quechua} & \textbf{Romanian} \\
\midrule
\multirow{3}{*}{Qwen2.5-VL-7B}
 & \gcell Base & \gcell 1.29 & \gcell 5.65 & \gcell 12.71 & \gcell 24.38 & \gcell 21.24 & \gcell 16.52 & \gcell 23.76 & \gcell 0.63 & \gcell 16.11 \\
 & \bcell VFA      & \bcell 1.39 & \bcell 6.52 & \bcell 12.79 & \bcell 25.27 & \bcell 19.76 & \bcell 14.69 & \bcell 22.81 & \bcell 0.23 & \bcell 15.94 \\
 & \textit{\small $\Delta$ Gains} & \gbf{+0.10} & \gbf{+0.87} & \gbf{+0.08} & \gbf{+0.89} & \rbf{--1.48} & \rbf{--1.83} & \rbf{--0.95} & \rbf{--0.40} & \rbf{--0.17} \\
 \midrule
\multirow{3}{*}{Idefics3-8B}
 & \gcell Base & \gcell 6.43 & \gcell 4.64 & \gcell 21.88 & \gcell 22.23 & \gcell 17.55 & \gcell 11.15 & \gcell 18.70 & \gcell 1.49 & \gcell 13.14 \\
 & \bcell VFA      & \bcell 6.11 & \bcell 6.61 & \bcell 16.76 & \bcell 25.51 & \bcell 20.29 & \bcell 16.30 & \bcell 24.02 & \bcell 1.56 & \bcell 17.18 \\
 & \textit{\small $\Delta$ Gains} & \rbf{--0.32} & \gbf{+1.97} & \rbf{--5.12} & \gbf{+3.28} & \gbf{+2.74} & \gbf{+5.15} & \gbf{+5.32} & \gbf{+0.07} & \gbf{+4.04} \\
 \midrule
\multirow{3}{*}{LLaVA-Next-8B}
 & \gcell Base & \gcell 3.91 & \gcell 0.00 & \gcell 0.00 & \gcell 0.14 & \gcell 3.36 & \gcell 0.62 & \gcell 0.39 & \gcell 1.90 & \gcell 0.47 \\
 & \bcell VFA      & \bcell 4.63 & \bcell 5.79 & \bcell 8.33 & \bcell 27.99 & \bcell 19.68 & \bcell 14.07 & \bcell 24.55 & \bcell 0.31 & \bcell 16.83 \\
 & \textit{\small $\Delta$ Gains} & \gbf{+0.72} & \gbf{+5.79} & \gbf{+8.33} & \gbf{+27.85} & \gbf{+16.32} & \gbf{+13.45} & \gbf{+24.16} & \rbf{--1.59} & \gbf{+16.36} \\

\midrule
\textbf{Model} & \textbf{Method} & \textbf{Russian} & \textbf{Swedish} & \textbf{Swahili} & \textbf{Telugu} & \textbf{Thai} & \textbf{Turkish} & \textbf{Ukrainian} & \textbf{Vietnamese} & \textbf{Chinese} \\

\midrule
\multirow{3}{*}{Qwen2.5-VL-7B}
 & \gcell Base & \gcell 23.06 & \gcell 23.87 & \gcell 9.70 & \gcell 4.74 & \gcell 0.54 & \gcell 13.34 & \gcell 14.72 & \gcell 28.61 & \gcell 6.97 \\
 & \bcell VFA      & \bcell 23.42 & \bcell 23.20 & \bcell 8.61 & \bcell 4.97 & \bcell 0.54 & \bcell 12.20 & \bcell 14.45 & \bcell 29.36 & \bcell 8.06 \\
 & \textit{\small $\Delta$ Gains} & \gbf{+0.36} & \rbf{--0.67} & \rbf{--1.09} & \gbf{+0.23} & \gbf{+0.00} & \rbf{--1.14} & \rbf{--0.27} & \gbf{+0.75} & \gbf{+1.09} \\
 \midrule
\multirow{3}{*}{Idefics3-8B}
 & \gcell Base & \gcell 14.62 & \gcell 20.33 & \gcell 10.47 & \gcell 6.08 & \gcell 1.19 & \gcell 13.65 & \gcell 9.31 & \gcell 22.23 & \gcell 4.21 \\
 & \bcell VFA      & \bcell 16.54 & \bcell 23.04 & \bcell 11.28 & \bcell 6.07 & \bcell 0.31 & \bcell 14.28 & \bcell 11.89 & \bcell 26.49 & \bcell 6.99 \\
 & \textit{\small $\Delta$ Gains} & \gbf{+1.92} & \gbf{+2.71} & \gbf{+0.81} & \rbf{--0.01} & \rbf{--0.88} & \gbf{+0.63} & \gbf{+2.58} & \gbf{+4.26} & \gbf{+2.78} \\
 \midrule
\multirow{3}{*}{LLaVA-Next-8B}
 & \gcell Base & \gcell 1.51 & \gcell 0.34 & \gcell 1.43 & \gcell 0.08 & \gcell 0.00 & \gcell 0.19 & \gcell 0.20 & \gcell 0.23 & \gcell 0.00 \\
 & \bcell VFA      & \bcell 15.42 & \bcell 21.89 & \bcell 9.56 & \bcell 2.76 & \bcell 0.00 & \bcell 10.32 & \bcell 12.44 & \bcell 18.99 & \bcell 4.43 \\
 & \textit{\small $\Delta$ Gains} & \gbf{+13.91} & \gbf{+21.55} & \gbf{+8.13} & \gbf{+2.68} & \gbf{+0.00} & \gbf{+10.13} & \gbf{+12.24} & \gbf{+18.76} & \gbf{+4.43} \\

\bottomrule
\end{tabular}%
}
\caption{Performance of different MLLMs on XM100 across multiple languages.}
\label{tab:breakdown_xm100}

\end{table*}

%% file: tables/tab_19_hi_ro_results.tex
\begin{table*}[h!]
    \centering
\small

    \setlength{\tabcolsep}{4pt}
    \setlength{\aboverulesep}{0pt}
    \setlength{\belowrulesep}{0pt}
    \renewcommand{\arraystretch}{1.2}

\setlength{\tabcolsep}{8pt}
\begin{tabular}{l|l| ccc}
\toprule
\textbf{Model} & \textbf{Method} & \textbf{MaXM (hi)} & \textbf{XMMU (hi)} & \textbf{MaXM (ro)} \\
\midrule

\multirow{3}{*}{\shortstack[l]{LLaVA-Next-8B}}
 & \gcell Base & \gcell 19.23 & \gcell 34.70 & \gcell 26.41 \\
 & \bcell VFA & \bcell 44.23 & \bcell 35.70 & \bcell 33.45 \\
 & \textit{\small $\Delta$ Gains} & \gbf{+25.00} & \gbf{+1.00} & \gbf{+7.04} \\ \hline

\multirow{3}{*}{\shortstack[l]{Idefics3-8B}}
 & \gcell Base & \gcell 62.31 & \gcell 42.30 & \gcell 37.68 \\
 & \bcell VFA & \bcell 65.77 & \bcell 46.00 & \bcell 44.01 \\
 & \textit{\small $\Delta$ Gains} & \gbf{+3.46} & \gbf{+3.70} & \gbf{+6.33} \\
\bottomrule
\end{tabular}
\caption{Multilingual performance of VFA on Hindi (hi) and Romanian (ro) benchmarks.}
\label{tab:multilingual_hi_ro_combined}

\end{table*}